\documentclass[11pt]{article}

\PassOptionsToPackage{table}{xcolor}
\usepackage[final]{acl}

\usepackage{times}
\usepackage{latexsym}

\usepackage[T1]{fontenc}
\usepackage[utf8]{inputenc}

\usepackage{microtype}

\usepackage{inconsolata}

\usepackage{graphicx}

\usepackage{booktabs}
\usepackage{amsmath}
\usepackage{amssymb}
\usepackage{multirow}

\usepackage{algorithm}
\usepackage{algpseudocode}
\usepackage{enumitem}

\usepackage{xcolor}
\usepackage{listings}
\usepackage{tcolorbox}
\tcbuselibrary{skins,breakable,listings}

\definecolor{PromptBlue}{HTML}{355C7D}
\definecolor{PromptBorder}{HTML}{CBD5E1}
\definecolor{PromptBg}{HTML}{FAFBFC}

\newtcblisting[
  auto counter,
  number within=section
]{promptbox}[2][]{
  enhanced,
  breakable,
  listing only,
  listing engine=listings,
  title={Prompt~\thetcbcounter: #2},
  colback=PromptBg,
  colframe=PromptBorder,
  colbacktitle=PromptBlue!9!white,
  coltitle=PromptBlue,
  fonttitle=\bfseries\small,
  boxrule=0.5pt,
  arc=1.5mm,
  left=3mm,
  right=3mm,
  top=2mm,
  bottom=2mm,
  before skip=10pt,
  after skip=10pt,
  listing options={
    language={},
    basicstyle=\ttfamily\small,
    columns=fullflexible,
    keepspaces=true,
    breaklines=true,
    breakatwhitespace=false,
    showstringspaces=false,
    numbers=none,
    aboveskip=0pt,
    belowskip=0pt
  },
  #1
}

\definecolor{ToolBrown}{HTML}{5D4037}
\definecolor{ToolBorder}{HTML}{D7CCC8}
\definecolor{ToolBg}{HTML}{FCFAF8}

\newtcblisting[
  auto counter,
  number within=section
]{toolbox}[2][]{
  enhanced,
  breakable,
  listing only,
  listing engine=listings,
  title={Tool~\thetcbcounter: #2},
  colback=ToolBg,
  colframe=ToolBorder,
  colbacktitle=ToolBrown,
  coltitle=white,
  fonttitle=\bfseries\small,
  boxrule=0.5pt,
  arc=1.5mm,
  left=3mm,
  right=3mm,
  top=2mm,
  bottom=2mm,
  before skip=10pt,
  after skip=10pt,
  listing options={
    language={},
    basicstyle=\ttfamily\footnotesize,
    columns=fullflexible,
    keepspaces=true,
    breaklines=true,
    breakatwhitespace=false,
    showstringspaces=false,
    numbers=none,
    aboveskip=0pt,
    belowskip=0pt
  },
  #1
}

\usepackage{array}
\usepackage{graphicx}

\newcommand{\lingan}[1]{{\color{black}#1}}

\title{\textsc{Uni-Mem}: Learning a Long-Horizon Agentic Memory Manager with Unified Online Policy Optimization}

\author{
  \textbf{Zhiqian Xia}\textsuperscript{1,3}\thanks{Equal contribution. Work done during internship.}
  \quad
  \textbf{Jielong Tang}\textsuperscript{2,3}\footnotemark[1]
  \quad
  \textbf{Ling-An Zeng}\textsuperscript{3}
  \\
  \textbf{Shixiong Kai}\textsuperscript{3}
  \quad
  \textbf{Rong Yao}\textsuperscript{3}
  \quad
  \textbf{Hao Kong}\textsuperscript{3}
  \quad
  \textbf{Xiao Dong}\textsuperscript{4}
  \\
  \textbf{Wei Wang}\textsuperscript{2}
  \quad
  \textbf{Mingxuan Yuan}\textsuperscript{3}
  \quad
  \textbf{Siyu Xia}\textsuperscript{1}
  \quad
  \textbf{Haifeng Xia}\textsuperscript{2}\thanks{Corresponding author.}
  \\[0.5ex]
  \textsuperscript{1}Southeast University  \\
  \textsuperscript{2}Sun Yat-sen University \\
  \textsuperscript{3}Huawei Technologies, Noah's Ark Lab \\
  \textsuperscript{4}Shandong Normal University
}

\begin{document}
\maketitle
\begin{abstract}
% Long-term agent memory is shifting from predefined workflows toward learnable policies for more adaptive and generalizable memory construction and retrieval. Existing learning-based methods either optimize construction alone or combine online retrieval reinforcement learning (RL) with offline construction supervision, limiting co-adaptation between the two stages. In this paper, we explore unified memory manager (Uni-Mem) learning via end-to-end online RL. However, 现有的RL算法 例如 GRPO joint online optimization of unified memory manager faces confounded group-relative rewards and sparse feedback over long memory-management trajectories. To this end, 我们提出Stage-Decoupled GRPO, xxx

Long-term agent memory is shifting from predefined workflows toward learnable policies for memory construction and retrieval. Existing learning-based methods either optimize construction alone or combine retrieval-side online reinforcement learning (RL) with offline construction supervision, limiting co-adaptation between the two stages. In this paper, we explore end-to-end online RL for a unified memory manager (\textsc{Uni-Mem}) that learns both capabilities under a shared policy. However, joint optimization with existing RL algorithms such as GRPO faces confounded group comparisons and sparse feedback over long memory-management trajectories. To this end, we propose Stage-Decoupled GRPO, which alternates construction and retrieval optimization while controlling variation from the other stage within each rollout group, yielding stage-specific learning signals. We further introduce Event-Wise Credit Assignment (ECA), which localizes feedback to responsible events and propagates event-level advantages to their constituent operations.
% ECA assesses memory sufficiency to distinguish construction from retrieval failures and incorporates historical memory versions to credit useful information overwritten by subsequent updates. 
% \todo{ECA assesses memory sufficiency to distinguish construction from retrieval failures and traces update lineages to credit \texttt{UPDATE} operations missed by event-level assignment.}
Experiments on LoCoMo and LongMemEval show that \textsc{Uni-Mem} outperforms the strongest RL-based memory, demonstrating the effectiveness of jointly learning memory construction and retrieval via Stage-Decoupled GRPO and ECA.

\end{abstract}

\section{Introduction}

\begin{figure}[t]
\centering
\includegraphics[width=1.0\linewidth]{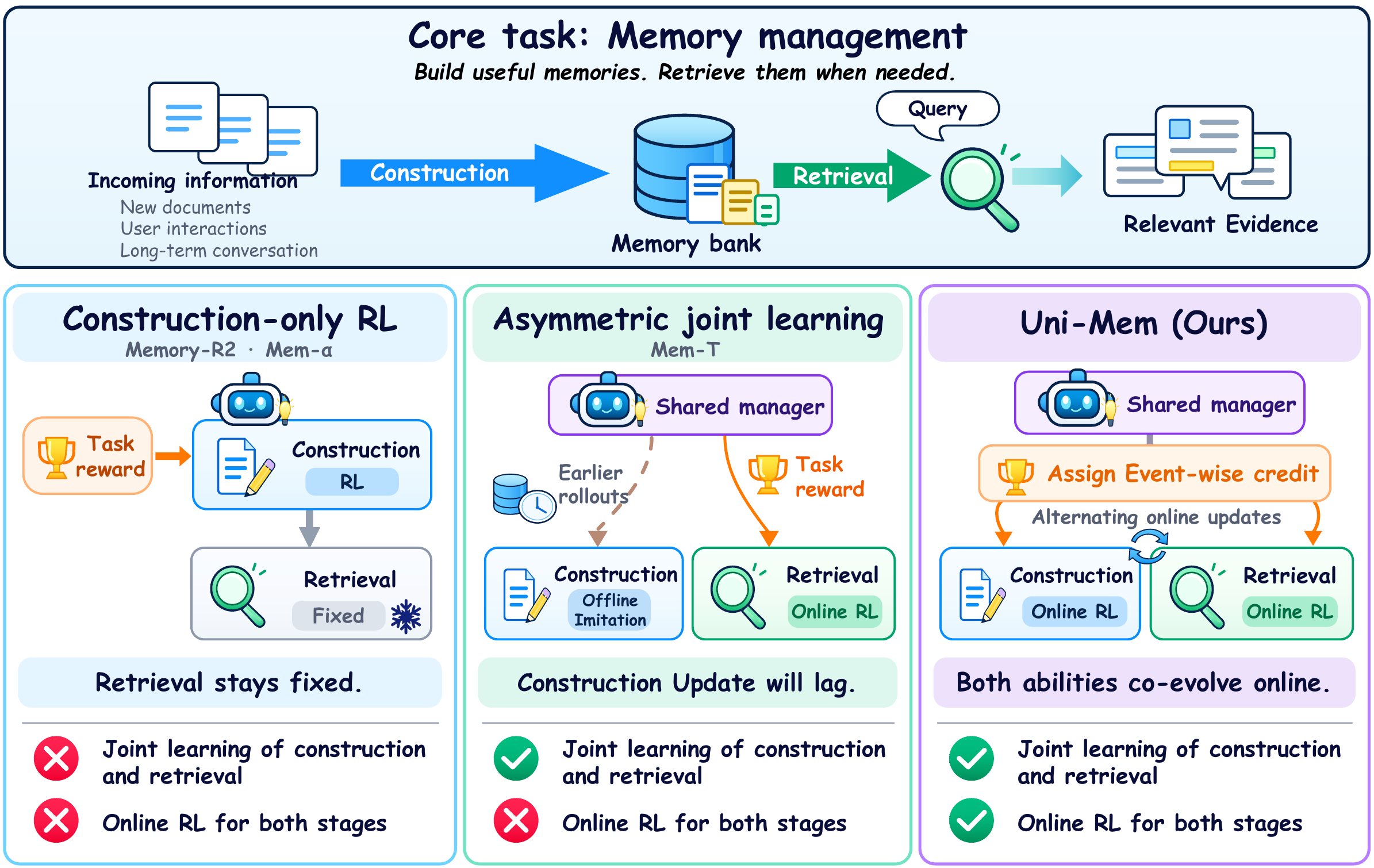}
\vspace{-10pt}
\caption{Comparison of training paradigms in different methods. Construction-only RL keeps retrieval fixed, while asymmetric joint learning combines offline construction imitation with online retrieval RL. \textsc{Uni-Mem} jointly optimizes both stages under a shared policy through alternating online RL and event-wise credit.}
\label{fig:intro}
\vspace{-12pt}
\end{figure}

% Large language models (LLMs) have evolved from basic text generators into agentic systems capable of autonomous reasoning and tool execution \citep{achiam2023gpt,yao2022react,schick2023toolformer}. However, their finite context window fundamentally bottlenecks internal state tracking, impairing long-term memory retention and cross-session goal consistency. To overcome this limitation, memory systems externalize past interactions for selective retrieval, thereby becoming a core component of long-horizon LLM agents \cite{wang2023augmenting,yu2026agentic}.

Long-term memory enables Large Language Model (LLM) agents to reuse information across interactions that exceed their context windows~\citep{wang2023augmenting, yu2026agentic}. Early memory systems, such as MemGPT~\citep{packer2023memgpt}, Mem0~\citep{chhikara2025mem0}, and A-Mem~\citep{xu2026mem}, rely on manually specified prompts and predefined workflows to organize and access past information. Despite their effectiveness, these systems remain constrained by frozen LLMs and rigid human priors, limiting the generalization of memory construction and retrieval to dynamic agent environments~\citep{xiong2026memory}.

To bridge this gap, recent works formulate memory management\footnote{In this paper, memory management encompasses construction (creation, update, and deletion) and retrieval.} as a learnable policy optimized via reinforcement learning (RL). 
% However, memory management may involve hundreds of construction and retrieval operations, making sparse outcome rewards alone insufficient for reliable process-level supervision in this long-horizon agentic training. 
Methods such as Memory-R2~\citep{yan2026memory_r2} and Mem-$\alpha$~\citep{wang2025mem} keep retrieval fixed and train construction with RL. 
% where the manager emits memory operations for each input in a single turn without conditioning on their execution results.
% Methods such as Memory-R2~\citep{yan2026memory_r1} and Mem-$\alpha$~\citep{wang2025mem} keep retrieval fixed and train construction operations with RL.
% , using QA-derived rewards for operation-level credit assignment. 
Such \textit{construction-only RL} learns how to build memory without jointly learning how to retrieve it adaptively, limiting memory utility in downstream tasks. To jointly learn both capabilities within a shared-parameter memory manager, approaches such as Mem-T~\citep{yue2026mem-t} combine online agentic RL of retrieval with offline imitation of hindsight-selected construction actions. Yet this \textit{asymmetric joint learning} can introduce a training lag: online retrieval RL continually updates the shared policy, whereas offline construction supervision remains anchored to earlier policy and memory states, hindering coordinated adaptation across the two stages. 
% This resulting mismatch between stale credit signals and the evolving policy-induced memory can bias manager optimization toward outdated advantage, hindering coordinated adaptation across the two stages.
% This resulting mismatch between stale credit signals and the evolving policy-induced memory distribution can bias construction optimization toward outdated utility estimates, hindering coordinated adaptation across the two stages. 
This motivates a central research question: 
\emph{How can a shared-parameter agentic memory manager, which operates through multi-turn tool interaction, jointly learn long-horizon construction and retrieval via end-to-end online RL?}
% \emph{How can a shared-parameter agentic memory manager jointly learn long-horizon construction and retrieval through end-to-end online RL?}

% , while keeping both stages aligned and assigning credit across them?}}

Achieving this goal poses two main challenges. \textbf{(1) Confounded group comparisons.}
Existing group-relative RL (e.g. GRPO~\cite{shao2024deepseekmath}) derives advantages from reward differences among sampled rollouts. For a unified memory manager, each rollout couples construction trajectories with a retrieval continuation, so its final reward reflects decisions from both stages. When both stages vary across rollouts, trajectory-level group-relative advantages can conflate their contributions, obscuring the learning signal for each stage.
% yield unreliable stage-specific learning signals.
% the performance of the stage being optimized.
% Memory construction precedes retrieval, yet online optimization of either stage can alter the behavior of the other under a shared-parameter setting. This coupling makes the end-to-end manager policy optimization non-stationary: credit signals from earlier construction rollouts may become misaligned with the subsequent retrieval rollouts, making it difficult to keep training feedback current and jointly preserve both capabilities.
% Meanwhile, the evolving construction policy changes the memory distribution encountered during retrieval, while the evolving retrieval policy changes the downstream utility of construction decisions. 
% This bidirectional dependence makes learning conditions non-stationary: credit signals from earlier construction rollouts may become misaligned with the 随后的 retrieval policy, 削弱了 feedback alignment and the coordinated maintenance of both capabilities.
\textbf{(2) Long-horizon credit attribution.}
% \lingan{In this setting, the manager processes an ordered stream of \textit{events}: each dialogue chunk triggers a construction event and each query a retrieval event. Beyond the already long event sequence, each event further unfolds into multiple tool calls, so a rollout spans hundreds of decisions while the outcome reward arrives only at answer time. Such sparse feedback neither localizes the query-relevant portions of a rollout nor reveals whether the required information was stored or successfully retrieved, and later updates may overwrite content produced by earlier decisions. Feedback must therefore be attributed to the relevant stage, events, and decisions within them.}
The manager policy performs hundreds of memory operations (tool calls) per rollout, yet receives the outcome reward only at answer time. Such sparse feedback neither localizes the query-relevant portions of a rollout nor reveals whether the required information was stored or successfully retrieved. Moreover, memory updates further complicate attribution by overwriting memory items produced by earlier decisions. Thus, feedback must be attributed to the relevant stage and operations within them. 

% Final outcome rewards depend on decisions across construction and retrieval trajectories that may span hundreds of operations. 
% % Such sparse outcome rewards alone thus provide insufficient guidance on which individual decisions contribute to the outcome, making reliable process supervision challenging in this long-horizon setting. 
% Such sparse feedback alone neither localizes the query-relevant portions of these trajectories nor reveals whether the required information was adequately stored or successfully retrieved. Subsequent updates further complicate attribution by overwriting memory content produced by earlier decisions.
% Effective policy optimization therefore requires attributing feedback to both the relevant stage and the specific construction or retrieval steps within it.
% Final outcome rewards jointly depend on earlier memory construction and subsequent retrieval, obscuring the contributions of individual operations. For example, an incorrect answer may arise because essential information was never retained or because retained information was not retrieved. 
% As both policies evolve, these causes become harder to disentangle. 
% Manager policy optimization must therefore attribute delayed, entangled outcome feedback to specific construction and retrieval decisions across long operation trajectories, providing reliable optimization signals for both stages.

% 这里先强调 在setting，一条连续轨迹 -> constr 和 retrieval，reward 是只有在回答问题后才能获取，如何将sparse outcome reward 分配到具体的阶段与operations

In this work, we propose \textbf{\textsc{Uni-Mem}}, a unified memory manager that jointly learns agentic construction and retrieval through end-to-end online RL. Fig.~\ref{fig:intro} compares the training paradigms of existing methods with ours. To address confounded group comparisons, we introduce \textbf{Stage-Decoupled GRPO}, which decouples the two stages contributions in group-relative advantages. Specifically, we alternate construction and retrieval optimization while controlling variation from the other stage within each rollout group. For construction, we bypass learned retrieval: since each memory record retains the identifier of its source chunk, we directly gather the records linked to a query's annotated supporting evidence (used only during training) and score them with a frozen answer model. This \textit{evidence-indexed score} reflects what was stored rather than how it was retrieved. For retrieval, all rollouts in a group share an identical memory state resampled from the current policy, so reward differences reflect retrieval alone. Alternating the two phases yields stage-specific signals, enabling both capabilities to co-evolve. 
% To address confounded comparisons, \textbf{Stage-Decoupled GRPO} alternates between construction and retrieval optimization while controlling variation from the other stage within each rollout group. For construction, we bypass learned retrieval: since each memory record retains the identifier of its source chunk, we directly gather the records linked to a query's annotated supporting evidence (used only during training) and score them with a frozen answer model. This \textit{evidence-indexed score} reflects what was stored rather than how it was retrieved. For retrieval, all rollouts in a group share an identical memory state resampled from the current policy, so reward differences reflect retrieval alone. Alternating the two phases yields stage-specific signals while letting both capabilities co-evolve.

To address long-horizon credit attribution, we introduce \textbf{Event-Wise Credit Assignment (ECA)}, which organizes multiple memory operations into \textit{events}\footnote{In this paper, we distinguish two types of events: construction event and retrieval event, denoted in Sec.~\ref{problem_formulation}}, triggered by an incoming dialogue chunk (\textit{construction event}) or a query (\textit{retrieval event}). 
% Each event serve as the unit of credit. 
ECA uses a query's supporting evidence to localize the responsible construction events, and reuses evidence-indexed scoring to penalize an unsuccessful retrieval event only when sufficient evidence has been stored. For update-robust evidence attribution, ECA tracks events along the same memory update chain and assigns them the corresponding credit. Since the event stream is fixed by the input data, rollouts share aligned event positions. ECA thus normalizes each event's feedback against same-position events across the rollout group, rather than at the trajectory level, and broadcasts the resulting advantage to all memory tool calls within that event. Our contributions are summarized as follows:

\begin{itemize}[leftmargin=\parindent]
    \item We propose \textsc{Uni-Mem}, a unified memory manager that jointly optimizes agentic construction and retrieval via end-to-end online RL, with \textbf{Stage-Decoupled GRPO} separating the two stages' contributions to group-relative signals.
    \item  We introduce \textbf{Event-Wise Credit Assignment}, which localizes outcome feedback to responsible events and computes advantages over position-aligned events across rollouts.
    % \item We propose \textsc{Uni-Mem}, a unified memory manager that optimizes agentic construction and retrieval through an end-to-end online RL, with \textbf{Stage-Decoupled GRPO} separating their contributions to group-relative learning signals.
    
    % \item We introduce \textbf{Event-Wise Credit Assignment} to localize feedback to relevant events and propagate event-level advantages to the corresponding decision steps.
    
    \item Extensive experiments on  LoCoMo and LongMemEval demonstrate that \textsc{Uni-Mem} outperforms the strongest RL-based memory manager, supporting the effectiveness of jointly learning memory construction and retrieval.
\end{itemize}

\section{Related Work}
\label{sec:related_work}

\paragraph{Long-term Memory Agents.}
Long-term memory enables LLM agents to retain and reuse information beyond their active context.
Retrieval-augmented generation (RAG) provides a basic mechanism for accessing external information~\citep{lewis2020retrieval}, while early agent-memory systems consolidate interactions into persistent records and reflections~\citep{park2023generative,zhong2024memorybank}.
Subsequent work emphasizes hierarchical storage and consolidation~\citep{packer2023memgpt,kang2025memory,fang2026lightmem}, or structured and associative organization~\citep{chhikara2025mem0,xu2026mem,rasmussen2025zep,gutierrez2025hipporag2, yuan2026metakgrag}.
Agentic retrieval further enables iterative evidence gathering from retained histories~\citep{yan2025gam}.
Complementing these architectural advances, \textsc{Uni-Mem} learns a shared policy for multi-turn memory construction and retrieval via adaptively tool calling, allowing both capabilities to adapt jointly to downstream tasks.

\paragraph{Reinforcement Learning for Memory Agents.}
RL-based memory methods optimize compressed working states, as in MEM1~\citep{zhou2026mem1} and MemAgent~\citep{yu2026memagent}, external-memory construction~\citep{wang2025mem}, or retrieval decisions~\citep{tan2025rmm,hsu2026horma}. AgeMem~\citep{yu2026agentic} coordinates long-term memory and short-term context.
Memory-R1~\citep{yan2026memory_r1} trains memory construction and a separate answer agent, whereas Memory-R2~\citep{yan2026memory_r2} extends this training to long horizons with local and global group-relative credit.
Denser construction supervision is also explored through intermediate rewards and evidence-based attribution~\citep{ma2026fine,shen2026membuilder}.
Beyond individual components, Mem-T~\citep{yue2026mem-t} combines online retrieval RL with offline imitation of construction actions selected using hindsight feedback. Different from existing approaches, \textsc{Uni-Mem} explores an end-to-end online RL of a shared-parameter manager policy while addressing the resulting challenges of confounded group comparisons and credit assignment with Decoupled GRPO and event-wise attribution. Appendix~\ref{app:related_comparison} further positions our work with representative methods' interaction patterns and training paradigm, including a detailed technical comparison with Mem-T.
% credit assignment when construction and retrieval are both learned online: Stage-Decoupled GRPO controls variation from the other stage within each rollout group, while Event-Wise Credit Assignment localize outcome feedback to event-aligned memory operations. Appendix~\ref{app:related_comparison} further position our work with representative methods' interaction patterns and training paradigm, including a detailed technical comparison with Mem-T.

% Appendix~\ref{app:comparison} further position our work with representative methods' interaction patterns and training paradigm, including a detailed comparison with Mem-T.
% \textsc{Uni-Mem} addresses credit assignment when construction and retrieval are both learned online: Stage-Decoupled GRPO controls variation from the other stage within each rollout group, while Event-Wise Credit Assignment distinguishes storage from retrieval failures, traces relevant memory updates, and propagates event-level credit to the corresponding operations.

\begin{figure*}[t]
\centering
\includegraphics[width=1.0\linewidth]{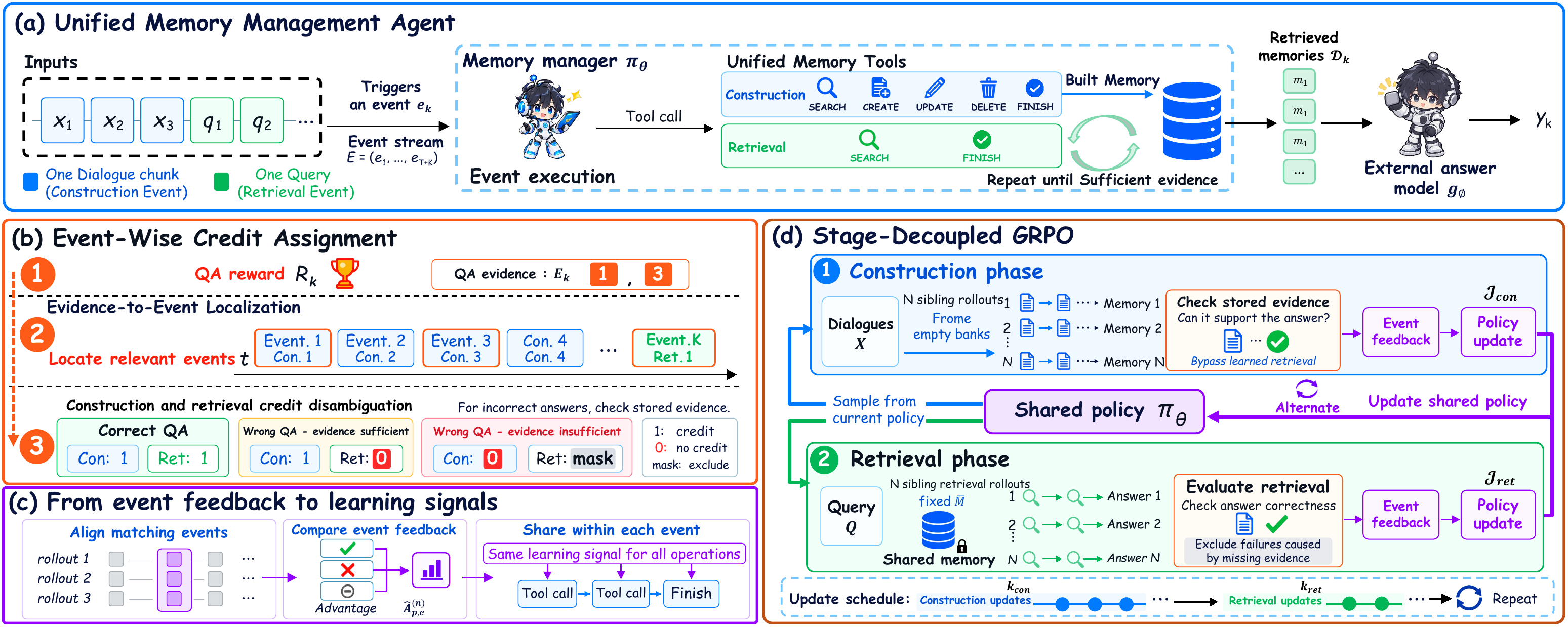}
% \vspace{-4pt}
\caption{Overview of \textsc{Uni-Mem}. (a) \textsc{Uni-Mem} unifies memory construction and retrieval through a shared tool-using agent, continually processing an ordered stream of events. (b) Event-Wise Credit Assignment attributes sparse outcome rewards to relevant events. (c) Event-level feedback is propagated to the corresponding memory operations (tool calls). (d) Stage-Decoupled GRPO alternates between construction and retrieval optimization.}
\label{fig:method}
\vspace{-12pt}
\end{figure*}

\section{Preliminaries}

\subsection{Problem Formulation}
\label{problem_formulation}

Our goal is to train a memory manager agent to maintain a
persistent external memory bank $\mathcal{M}$ and retrieve
relevant memories for downstream question answering.
We formulate this task as the continual processing of an
ordered stream of events.

Let $X=(x_1,\ldots,x_T)$ and $Q=(q_1,\ldots,q_K)$ denote
the sequences of dialogue chunks and queries, respectively.
A \emph{construction event} $e_t^{\mathrm{con}}$ is triggered
by a dialogue chunk $x_t$ and requires the agent to decide
whether and how to revise the memory bank.
A \emph{retrieval event} $e_k^{\mathrm{ret}}$ is triggered
by a query $q_k$ and requires the agent to gather supporting
evidence $\mathcal{D}_k\subseteq\mathcal{M}$.
The arrival order of these inputs defines the event stream:
\begin{equation}
\mathcal{E}=(e_1,\ldots,e_{T+K}).
\end{equation}
In our datasets, all queries arrive after the dialogue, so
retrieval operates on the bank built from $X$.

For each query, a frozen external answer model $g_\phi$
responds using the retrieved evidence:
\begin{equation}
y_k=g_\phi(q_k,\mathcal{D}_k).
\label{eq:answer_generation}
\end{equation}
We aim to jointly optimize the agent's construction and
retrieval behavior under a shared policy to improve
downstream answer accuracy.

\subsection{Unified Memory Management Agent}

We instantiate the memory manager as a tool-using agent
governed by a shared policy $\pi_\theta$, as illustrated
in Fig.~\ref{fig:method} (a).
The memory bank stores records of the form
$m_i=(i,c_i,\mathrm{content}_i)$, where $i$ is a unique
record ID, $c_i\in\{1,\ldots,T\}$ is the index of the dialogue chunk
from which the record was created,
and $\mathrm{content}_i$ is the stored textual content.

\textit{Both event types support multiple rounds of tool interaction}.
At each step, the agent selects a tool call based on the
event type, the triggering input, and the preceding tool
calls and observations within the event.
It then executes the call and uses the returned result
to guide its next decision.

During construction, the agent can use \texttt{SEARCH},
\texttt{CREATE}, \texttt{UPDATE}, and \texttt{DELETE}
to inspect and revise the bank.
During retrieval, it uses \texttt{SEARCH} to gather evidence,
potentially conducting successive searches based on
earlier results.
For both event types, the agent terminates by calling
\texttt{FINISH}, which also submits the selected evidence
$\mathcal{D}_k$ for retrieval events. Details of the memory
representation and tool specifications are provided in
Appendices~\ref{app:memory_representation}
and~\lingan{\ref{app:tool_specifications}}, respectively.

\section{Method}

In this section, we present the training framework of
\textsc{Uni-Mem}, as shown in Fig.~\ref{fig:method}. Online RL alternates between a construction phase
and a retrieval phase (Section~\ref{sec:sd_grpo}). We first describe
Event-Wise Credit Assignment (Section~\ref{sec:method_eca}), which
assigns rewards to events in each phase, and then the optimization
procedure, which combines an SFT cold start with Stage-Decoupled GRPO.

\subsection{Event-Wise Credit Assignment}
\label{sec:method_eca}

In long-horizon memory management, the outcome reward arrives only
after long sequences of tool calls, making it hard to
identify which tool calls contributed. We therefore
introduce Event-Wise Credit Assignment (ECA), which localizes
each query's reward to the events responsible for it.

Let $y_k^\star$ denote the reference answer to query $q_k$. A fixed
LLM judge $\mathcal{J}_\psi$ evaluates the generated response $y_k$
and assigns the task reward
\begin{equation}
R_k=\mathcal{J}_\psi(q_k,y_k,y_k^\star)\in\{0,1\}.
\end{equation}

\paragraph{Evidence-to-event localization.}
Let $E_k\subseteq\{1,\ldots,T\}$ denote the indices of the
dialogue chunks that contain the annotated evidence for $q_k$, which
are used only during training. The construction events related to
$q_k$ are
\begin{equation}
\mathcal{E}_k^{\mathrm{con}}
=\{e_t^{\mathrm{con}}\mid t\in E_k\}.
\label{eq:evidence_event_localization}
\end{equation}

\paragraph{Evidence-indexed memory scoring.}
An incorrect answer may result from insufficient stored information
or from unsuccessful retrieval. To distinguish the two, for a memory
bank $\mathcal{B}$, $\operatorname{Index}(E_k,\mathcal{B})$ selects
the current versions of the records written in
$\mathcal{E}_k^{\mathrm{con}}$, and the frozen answer model answers
$q_k$ from these records:
\begin{equation}
\begin{aligned}
\widetilde{y}_k(\mathcal{B})
&=g_\phi\!\left(q_k,\operatorname{Index}(E_k,\mathcal{B})\right),\\
\mathcal{S}_k(\mathcal{B})
&=\mathcal{J}_\psi\!\left(q_k,\widetilde{y}_k(\mathcal{B}),y_k^\star\right)
\in\{0,1\}.
\end{aligned}
\label{eq:memory_scoring}
\end{equation}
Thus, $\mathcal{S}_k(\mathcal{B})$ serves as a proxy for whether the stored
information suffices to answer $q_k$, without involving the learned
retrieval policy.

\paragraph{Event-level credit.}
In the construction phase, rollout $n$ produces a final bank
$\mathcal{B}^{(n)}$, and each such event accumulates the
scores of its related queries:
\begin{equation}
\bar{R}_{\mathrm{con},e}^{(n)}
=\sum_{k:\,e\in\mathcal{E}_k^{\mathrm{con}}}
\mathcal{S}_k(\mathcal{B}^{(n)}).
\label{eq:construction_credit}
\end{equation}
Note that events that are not related to any query in a rollout receive no
return and are excluded from the group at their position.
In the retrieval phase, all rollouts share a bank
$\overline{\mathcal{M}}$, and the retrieval event triggered by $q_k$
receives
\begin{equation}
\bar{R}_{\mathrm{ret},e_k}^{(n)}
=\begin{cases}
R_k^{(n)}, & R_k^{(n)}\!=\!1\lor\mathcal{S}_k(\overline{\mathcal{M}})\!=\!1,\\
\bot, & \text{otherwise},
\end{cases}
\label{eq:event_credit}
\end{equation}
where $\bot$ marks an excluded event. A failed retrieval is thus
penalized only when the shared bank supports a correct answer;
otherwise, the failure is attributed to construction, and the event
contributes neither to group statistics nor to policy gradients.
The resulting returns are converted into event-level advantages in
Section~\ref{sec:sd_grpo}.

% \paragraph{Update-robust evidence attribution.}
% Records written in $\mathcal{E}_k^{\mathrm{con}}$ may later be
% updated or deleted by events outside $\mathcal{E}_k^{\mathrm{con}}$,
% which then receive no credit. As a minor refinement, we track the
% update lineage of these records during training and add such events
% to $\mathcal{E}_k^{\mathrm{con}}$, so that they share the same score
% $\mathcal{S}_k(\mathcal{B}^{(n)})$. This only enlarges the set of
% credited events and leaves Eqs.~\eqref{eq:memory_scoring}
% and~\eqref{eq:construction_credit} unchanged. Details are provided
% in Appendix~\ref{app:credit_computation}.

\lingan{
\paragraph{Update-robust evidence attribution.}
Records written in $\mathcal{E}_k^{\mathrm{con}}$ may later be
updated or deleted by events outside $\mathcal{E}_k^{\mathrm{con}}$,
which then receive no credit. We therefore track the update lineage
of these records during training and form a rollout-specific credit
set $\widetilde{\mathcal{E}}_k^{\mathrm{con},(n)}\supseteq\mathcal{E}_k^{\mathrm{con}}$
that additionally contains such events, which share the same score
$\mathcal{S}_k(\mathcal{B}^{(n)})$. With this attribution,
$\widetilde{\mathcal{E}}_k^{\mathrm{con},(n)}$ replaces
$\mathcal{E}_k^{\mathrm{con}}$ in Eq.~\eqref{eq:construction_credit},
while memory selection in Eq.~\eqref{eq:memory_scoring} still uses
the original $E_k$. See Appendix~\ref{app:credit_computation}.
}

\subsection{Unified Manager Policy Optimization}

In this section, we describe the optimization procedure for
our \textsc{Uni-Mem}. We first cold-start the shared
policy through supervised fine-tuning (SFT) on demonstration
data, then further optimize it with Stage-Decoupled GRPO, as shown in Fig.~\ref{fig:method}(d).

\paragraph{Supervised cold start.}
To establish basic construction and retrieval capabilities, we
initialize the shared policy through supervised fine-tuning on memory
management demonstrations.
Details of the demonstration data and training objective
are provided in Appendix~\ref{app:supervised_cold_start}.

\paragraph{Stage-Decoupled GRPO.}
\label{sec:sd_grpo}
Directly applying GRPO to end-to-end rollouts can conflate
construction and retrieval contributions when both stages
vary, obscuring the learning signal for each stage.
To address this, we introduce Stage-Decoupled GRPO, which
alternates construction and retrieval optimization while
controlling variation from the other stage within each
rollout group. Construction rollouts are evaluated without
learned retrieval, while retrieval rollouts share a fixed
memory bank. Both phases update the same shared manager
$\pi_\theta$.

At each iteration, a snapshot $\pi_{\theta_{\mathrm{old}}}$
of the current manager samples $N$ sibling rollouts per group
for the scheduled phase. Construction rollouts process the
same dialogues from empty banks; retrieval rollouts for each
query share a bank $\overline{\mathcal{M}}$ that this snapshot
constructs once by processing the dialogue before retrieval.
Algorithm~\ref{alg:phase_reward_assignment} assigns
event-level returns $\bar{R}_{p,e}^{(n)}$, which are used to update
$\pi_\theta$.

\begin{algorithm}[t]
\caption{Phase-Specific Event Reward Assignment}
\label{alg:phase_reward_assignment}
\small
\algrenewcommand{\algorithmicrequire}{\textbf{Input:}}
\algrenewcommand{\algorithmicensure}{\textbf{Output:}}
\begin{algorithmic}[1]
\Require Phase $p$; $N$ sampled rollouts with their memory banks;
query annotations $\{q_k,E_k,y_k^\star\}_{k=1}^{K}$
\Ensure Event returns $\{\bar{R}_{p,e}^{(n)}\}$, with $\bot$ marking
excluded events
\If{$p=\mathrm{con}$}
    \For{$n=1,\ldots,N$}
        \State Form $\widetilde{\mathcal{E}}_k^{\mathrm{con},(n)}$ from
        Eq.~\eqref{eq:evidence_event_localization} and the update
        lineage, for all $k$
        \State $s_k^{(n)}\gets\mathcal{S}_k(\mathcal{B}^{(n)})$,
        for all $k$
        \State $\bar{R}_{\mathrm{con},e}^{(n)}\gets
        \sum_{k:\,e\in\widetilde{\mathcal{E}}_k^{\mathrm{con},(n)}}s_k^{(n)}$,
        for related events $e$
    \EndFor
\Else
    \For{$k=1,\ldots,K$}
        \State Evaluate task rewards $\{R_k^{(n)}\}_{n=1}^{N}$
        \If{$\exists n:R_k^{(n)}=0$}
            \State Evaluate $\mathcal{S}_k(\overline{\mathcal{M}})$
            on the shared bank
        \EndIf
        \State Assign $\bar{R}_{\mathrm{ret},e_k}^{(n)}$ by
        Eq.~\eqref{eq:event_credit}, for all $n$
    \EndFor
\EndIf
\State \Return $\{\bar{R}_{p,e}^{(n)}\}$
\end{algorithmic}
\end{algorithm}

Using the assigned returns, we group corresponding events across
rollouts by their triggering dialogue-chunk or query ID. Since all
rollouts of a training sample process the same stream of dialogue
chunks and queries, events triggered by the same input are aligned
across rollouts, even though the tool calls performed within
them may differ. Differences in the preceding memory states are
treated as part of the sampled trajectory, as in standard GRPO.
Construction events without a return and retrieval events marked
$\bot$ in Eq.~\eqref{eq:event_credit} are excluded from both group
statistics and policy optimization. Groups with fewer than two valid
returns or zero variance are discarded.
For each retained group in phase
$p\in\{\mathrm{con},\mathrm{ret}\}$, we compute
\begin{equation}
\widehat{A}_{p,e}^{(n)}
=
\frac{\bar{R}_{p,e}^{(n)}-\mu_{p,e}}
{\sigma_{p,e}+\delta},
\label{eq:event_advantage}
\end{equation}
where $\mu_{p,e}$ and $\sigma_{p,e}$ are the mean and
standard deviation of the valid returns,
and $\delta>0$ is a numerical stabilizer. Under event-level credit
assignment, all tool calls within an event share its advantage, as illustrated in Fig.~\ref{fig:method}(c).

Following GRPO, every token of a retained tool call is
optimized with the clipped surrogate, using the advantage of its
event:
\begin{equation}
\begin{split}
\mathcal{J}_p(\theta)=\mathbb{E}_{n}\Big[
&\tfrac{1}{|\tau_p^{(n)}|}\textstyle\sum_{o\in\tau_p^{(n)}}
\min\big(\rho_o\widehat{A}_{p,e(o)}^{(n)},\\
&\operatorname{clip}(\rho_o,1\!-\!\epsilon,1\!+\!\epsilon)\,
\widehat{A}_{p,e(o)}^{(n)}\big)\Big],
\end{split}
\label{eq:phase_objective}
\end{equation}
where $\tau_p^{(n)}$ contains the tokens of the tool calls in rollout
$n$ retained in phase $p$, $\rho_o$ is the token-level likelihood ratio
between $\pi_\theta$ and $\pi_{\theta_{\mathrm{old}}}$, and $e(o)$ is
the event containing token $o$. Rollouts without retained tokens are
skipped; no KL penalty is used.

The two phases alternate according to
$\boldsymbol{\kappa}=(\kappa_{\mathrm{con}},\kappa_{\mathrm{ret}})$:
$\kappa_{\mathrm{con}}$ consecutive construction updates are followed
by $\kappa_{\mathrm{ret}}$ retrieval updates (denoted $C$ and $R$ in
Figure~\ref{fig:further_analysis}). The RL objective is
\begin{equation}
\mathcal{J}_{\mathrm{RL}}(\theta)
=\lambda_{\mathrm{con}}\mathcal{J}_{\mathrm{con}}(\theta)
+\lambda_{\mathrm{ret}}\mathcal{J}_{\mathrm{ret}}(\theta),
\label{eq:alternating_rl_objective}
\end{equation}
where $\lambda_{\mathrm{con}}=1$ and $\lambda_{\mathrm{ret}}=0$
during construction updates, and vice versa during retrieval updates.

\section{Experiments}

\subsection{Experiment Setup}
\textbf{Datasets and Evaluation Metrics.}
We evaluate our framework on LoCoMo~\cite{locomo2024evaluating} and LongMemEval~\cite{wu2024longmemeval}.
% , and HotpotQA~\cite{yang2018hotpotqa}. 
For a fair comparison, we follow the training data configuration of Memory-R2~\cite{yan2026memory_r2}, partitioning LoCoMo into RL training, validation, and test sets with a 2:1:7 split. We use LongMemEval for out-of-distribution evaluation. We report LLM-as-a-Judge scores ($J$) using the official Mem0~\citep{chhikara2025mem0} judge prompt.

\noindent \textbf{Baselines.} We compare \textsc{Uni-Mem} against two groups of baselines.
\textit{(I) Training-free methods} include RAG for conversation retrieval, as well as A-Mem~\citep{xu2026mem}, Mem0~\citep{chhikara2025mem0}, and MemoryOS~\citep{kang2025memory}.
\textit{(II) Training-based methods} include MEM1~\citep{zhou2026mem1}, MemAgent~\citep{yu2026memagent}, Memory-R1~\citep{yan2026memory_r1}, and Memory-R2~\citep{yan2026memory_r2}, and Mem-T~\citep{yue2026mem-t}. These methods typically use RL to optimize memory phase. For a fair comparison, we re-evaluate the Memory-R2 (OSS) variant, which uses GPT-OSS-120B~\cite{agarwal2025gpt} as its answer model. Details of baselines are in Appendix~\ref{app:baseline}

\noindent \textbf{Implementation Details.}
We use Qwen3-4B-Instruct~\cite{yang2025qwen3} as the backbone of our memory manager. GPT-OSS-120B serves as the external answer model, and GPT-4o-mini~\cite{hurst2024gpt} serves as the judge model. Additional training details are provided in Appendix~\ref{app:training_details}.

\subsection{Main Results}

\begin{table}[t]
\centering
\footnotesize
\setlength{\tabcolsep}{2.5pt}
\renewcommand{\arraystretch}{1.05}
\begin{tabular}{@{}lccccc@{}}
\toprule
\textbf{Model}
& \textbf{Multi}
& \textbf{Open}
& \textbf{Single}
& \textbf{Temp.}
& \textbf{Overall} \\
\midrule
\rowcolor{gray!30}
\multicolumn{6}{l}{\textit{Training-free Methods}} \\
RAG            & 15.06 & 19.28 & 12.79 & 5.43  & 12.17 \\
A-MEM      & 40.78 & 31.32 & 46.90 & 28.68 & 40.78 \\
Mem0        & 61.92 & 48.19 & 65.20 & 38.76 & 53.30 \\
MemoryOS    & 48.12 & 38.55 & 63.14 & 24.81 & 51.26 \\
\midrule
\rowcolor{gray!30}
\multicolumn{6}{l}{\textit{Training-based methods}} \\
MEM1       & 41.29 & 43.10 & 57.22 & 44.73 & 50.78 \\
MemAgent      & 70.65 & 62.07 & 78.61 & 56.96 & 71.52 \\
Memory-R1  & 62.34 & 40.96 & 67.81 & 49.61 & 61.51 \\
% Memory-R2 (OSS)$^{\dagger}$ & 85.07 & 74.14 & 90.58 & 83.33 & 87.10 \\
% Memory-R2$^{\dagger}$       & 80.93 & 67.53 & 86.80 & 69.90 & 80.99 \\
Memory-R2$^{\dagger}$       & 69.15 & 61.29 & 78.61 & 57.38 & 71.26 \\
Memory-R2 (OSS)$^{\dagger}$ & 74.13 & 70.97 & 81.15 & 74.68 & {77.87} \\
Mem-T$^{\dagger}$           & 68.16 & 58.06 & 81.15 & \underline{76.79} & 76.49 \\
\midrule
% Uni-Mem(w/o training) & 78.61 & 70.97  & 84.04  &  57.81&76.58 \\
\textsc{Uni-Mem}(w/o sft)     &  \underline{81.09} &  \underline{75.81} & \underline{88.79}  & 62.87 & \underline{80.99} \\
\textsc{Uni-Mem}            & \textbf{87.06}  &\textbf{79.03}  & \textbf{91.51}  & \textbf{86.08} &\textbf{88.80}  \\
\bottomrule
\end{tabular}
\caption{
% Performance comparison on LoCoMo using LLM-as-a-Judge scores. Bold and underlined results indicate the best and second-best performance, respectively. $^{\dagger}$ marks results from our re-evaluation of the officially released Memory-R2 and Mem-T checkpoints with the Mem0 official judge prompt. Results for all other baselines are from~\cite{yan2026memory_r2}.
\lingan{
Performance comparison on LoCoMo (LLM-as-a-Judge, \%). Bold and underlined results indicate the best and second-best, respectively. $^{\dagger}$ marks our re-evaluation of the officially released checkpoints under the same protocol as \textsc{Uni-Mem}. Other baseline results are taken from \citet{yan2026memory_r2} under their original protocol, as reference only (Appendix~\ref{app:memory_r2_reevaluation}).
}
\vspace{-0.2cm}
}
\label{tab:performance_comparison}
\end{table}

% As shown in Table~\ref{tab:result_on_locomo}, Uni-Mem achieves the best performance across all four question types, with an overall $J$ score of 88.80\%.
% Under matched evaluation conditions, it surpasses Memory-R2 (OSS)$^{\ddagger}$ and Mem-T$^{\ddagger}$ by 10.93 and 12.31 percentage points, respectively.
% Its gains over the best re-evaluated baseline scores on multi-hop and temporal questions are 12.93 and 9.29 points, respectively.
% Even without additional training, our framework achieves 76.58\% with the base Qwen3-4B, outperforming all training-free baselines and most trained methods, supporting the effectiveness of its agentic memory-management design.
% Building on this strong starting point, direct RL improves the score to 80.99\%, while SFT initialization followed by RL yields the best performance.
% These results support jointly optimizing memory construction and retrieval through Stage-Decoupled GRPO and Event-Wise Credit Assignment, which provide stage-specific learning signals and localized supervision for memory operations.

\textbf{Superiority over existing baselines.}
As shown in Table~\ref{tab:performance_comparison}, \textsc{Uni-Mem} achieves the best performance across all four question types, with an overall $J$ score of 88.80\%.
It surpasses the strongest baseline, Memory-R2 (OSS), by 10.93 percentage points and Mem-T by 12.31 points, with gains of 12.93 and 9.29 points over the strongest competing results on multi-hop and temporal questions, respectively.
% Even without additional training, our framework achieves 76.58\% with the base Qwen3-4B, outperforming all training-free baselines and most trained methods and highlighting the effectiveness of its agentic memory-management design.
% 我们进一步移除SFT，
Additionally, \textsc{Uni-Mem} still achieves 80.99\% without SFT, outperforming the strongest baseline by 3.12\%. 
These results support the effectiveness of jointly optimizing memory construction and retrieval through Stage-Decoupled GRPO and Event-Wise Credit Assignment, which provide stage-specific learning signals and localized supervision for memory operations.

\begin{figure}[t]
    \centering
    \includegraphics[width=\columnwidth]{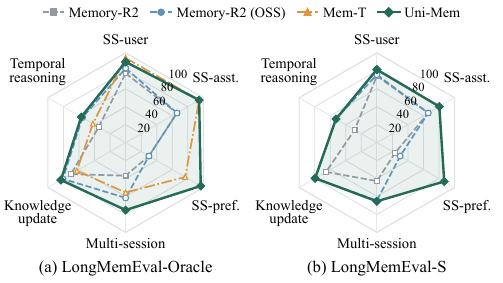}
    \caption{OOD accuracy ($J$, \%) on LongMemEval-Oracle
    and LongMemEval-S. SS: single-session; asst.: assistant;
    pref.: preference. Both panels use a 0--100\% scale.
    Mem-T results are shown only for Oracle.}
    \vspace{-0.2cm}
    \label{fig:ood_generalization}
\end{figure}

\noindent \textbf{Out-of-Distribution Generalization.}
To assess whether the memory-management capabilities learned
on LoCoMo transfer to a different benchmark, we evaluate
\textsc{Uni-Mem} on LongMemEval-Oracle and
LongMemEval-S~\citep{wu2024longmemeval}. 
Figure~\ref{fig:ood_generalization} presents the category-level results. On LongMemEval-Oracle, \textsc{Uni-Mem} substantially outperforms all baselines on single-session-preference and multi-session questions. It also achieves notable gains in knowledge update and temporal reasoning over Mem-T, a prior unified memory manager. On the larger-scale LongMemEval-S, \textsc{Uni-Mem} consistently outperforms Memory-R2 and its variants across all six task categories. These results demonstrate the transferability of its learned memory-management capabilities beyond the LoCoMo training distribution.

\begin{table}[t]
\centering
\footnotesize
\setlength{\tabcolsep}{2.5pt}
\renewcommand{\arraystretch}{1.05}
\begin{tabular}{@{}lccccc@{}}
\toprule
\textbf{Model}
& \textbf{Multi}
& \textbf{Open}
& \textbf{Single}
& \textbf{Temp.}
& \textbf{Overall} \\
\midrule
\rowcolor{gray!30} \multicolumn{6}{l}{\textit{Qwen3-4B-Instruct}} \\
$M_0$    & 78.61 & 70.97  & 84.04  &  57.81&76.58 \\
$M_1$     &  81.09 & 69.35  & 85.57 & 54.01 &76.95 \\
% M1       &  83.08 & 75.81 & 86.08  & 54.43&78.05 \\
$M_2$       & 82.59& 72.58 & 86.59 & 55.27 & 78.24\\
$M_3$    & 84.08 &  66.13 &  87.10&  65.82 &80.72 \\
$M_4$    &  81.09 &  75.81 & 88.79  & 62.87 &80.99\\
\midrule 
\rowcolor{gray!30} \multicolumn{6}{l}{\textit{Qwen3-4B-Instruct after SFT}} \\
$M_0$          &82.09 & 69.35 & 88.12 &  77.22 &83.56 \\
$M_1$          & 84.08 & 70.97  & 88.62 &  75.95 &84.02 \\
% M1      & 86.07 & 75.81  &90.83 & 78.06 &86.32 \\
$M_2$        & 84.08 &  74.19  & 90.66&83.97  &87.05\\
$M_3$    & 88.56  &  77.42& 90.32  & 84.81 &88.06\\
$M_4$  & 87.06  &79.03  & 91.51  & 86.08 & 88.80 \\
\bottomrule
\end{tabular}
\caption{
Ablation study on LoCoMo ($J$, \%) with raw or SFT-initialized Qwen3-4B-Instruct as start point ($M_0$).
}
\label{tab:ablation}
\end{table}

\subsection{Ablation Study}

\textbf{Ablation setting.} To validate the effectiveness of \textsc{Uni-Mem}’s module
designs, we conduct an incremental ablation study. Specifically, $M_0$ serves as the starting point, using either the original or SFT-initialized base manager. $M_1$ jointly optimizes memory construction and retrieval using standard GRPO. $M_2$ adds Event-Wise Credit Assignment to $M_1$. $M_3$ replaces standard GRPO in $M_2$ with Stage-Decoupled GRPO. Finally, $M_4$ incorporates update-robust evidence attribution into $M_3$, yielding the full \textsc{Uni-Mem} framework. Table~\ref{tab:ablation} presents the ablation results using either the original or SFT-initialized Qwen3-4B-Instruct model. The single-stage rollout and reward-assignment procedures for
$M_1$ and $M_2$ are detailed in Appendix~\ref{app:ablation_configurations}.

\noindent \textbf{Event-wise Credit Assignment.}
Sparse outcome rewards offer limited guidance for credit assignment across hundreds of memory operations. Adding ECA ($M_1\rightarrow M_2$) improves overall $J$ by 1.29 and 3.03 percentage points without and with SFT, respectively. These gains support the effectiveness of ECA, which localizes feedback to relevant events and their operations. Update-robust attribution ($M_3\rightarrow M_4$) continually improves the performance, supporting the additional benefit of credit tracing along memory update lineages.

\noindent \textbf{Stage-Decoupled GRPO.}
 $M_0\rightarrow M_1$ brings very limited gain (only 0.37\%), which indicates that group-relative advantages can conflate construction and retrieval effects. 
 % Stage-Decoupled GRPO ($M_2\rightarrow M_3$) improves overall $J$ by 2.48\%. In particular, temporal accuracy rises by 10.55 points, and multi-hop accuracy increases by 4.48 points after SFT. 
 Stage-Decoupled GRPO ($M_2\rightarrow M_3$) improves overall $J$ by 2.48 percentage points without SFT. In particular, temporal accuracy rises by 10.55 points in this setting, and multi-hop accuracy increases by 4.48 points after SFT.
 These gains support the effectiveness of alternating optimization and controlling cross-stage variation in Stage-Decoupled GRPO.

% \textbf{Event-wise Credit Assignment.}
% Sparse outcome rewards provide limited guidance on which construction or retrieval operations deserve credit in trajectories spanning hundreds of decisions. Adding ECA (M1$\rightarrow$M2) improves overall $J$ by 1.29 and 3.03 percentage points without and with SFT initialization, respectively. With SFT, temporal accuracy increases by 8.02 points. These gains support localizing feedback to relevant events and their constituent operations while distinguishing storage failures from retrieval failures. Incorporating update-robust attribution (M3$\rightarrow$M4) further improves overall $J$ by 0.27 and 0.74 points, respectively, supporting the additional benefit of tracing credit along memory update lineages.

% \textbf{Stage-Decoupled GRPO.}
% Although ECA localizes feedback, group-relative advantages can still mix construction and retrieval effects when both stages vary across rollouts. Replacing standard GRPO with Stage-Decoupled GRPO (M2$\rightarrow$M3) improves overall $J$ by 2.48 and 1.01 percentage points without and with SFT initialization, respectively. Notable gains include 10.55 points on temporal questions without SFT and 4.48 points on multi-hop questions with SFT. With ECA already enabled in M2, the additional gains from M3 support the effectiveness of controlling cross-stage variation in group-relative policy optimization.

\subsection{Further Analysis}

\begin{table}[t]
\centering
\footnotesize
\setlength{\tabcolsep}{2.5pt}
\renewcommand{\arraystretch}{1.05}
\begin{tabular}{@{}cc|ccccc@{}}
\toprule
\textbf{Memory}
& \textbf{Search}
& \textbf{Multi}
& \textbf{Open}
& \textbf{Single}
& \textbf{Temp.}
& \textbf{Overall} \\
\midrule
$P_0$ & $Q_0$ & 78.61 & 70.97  & 84.04  &  57.81&76.58 \\
$P_0$ & $Q_+$ & 86.07& 67.74&88.29 &62.45 &81.08 \\
$P_+$ & $Q_0$ &77.61 &69.35&89.98 &81.86 &84.76 \\
$P_+$ & $Q_+$ & 87.06  &79.03  & 91.51  & 86.08 &88.80 \\
\bottomrule
\end{tabular}
\caption{
Evaluation of memory construction and retrieval capabilities on LoCoMo ($J$, \%) by cross-combining memory banks and retrieval policies.
$P_0$ and $P_+$ denote memory banks constructed by the base Qwen3-4B manager and the trained \textsc{Uni-Mem} manager, respectively;
$Q_0$ and $Q_+$ denote their corresponding retrieval policies.
}
\label{tab:construction_search}
\end{table}

\textbf{Evaluating Construction and Retrieval Capabilities.}
We analyze the memory construction and retrieval capabilities of \textsc{Uni-Mem}.
To assess the improvements in each capability, we cross-combine memory banks and retrieval policies from the base and trained managers, as shown in Table~\ref{tab:construction_search}.
With the base retrieval policy fixed, replacing $P_0$ with $P_+$ improves the overall $J$ score by 8.18 percentage points, demonstrating the increased downstream utility of the constructed memories.
Conversely, replacing $Q_0$ with $Q_+$ improves the overall score by 4.50 and 4.04 points on $P_0$ and $P_+$, respectively, indicating more effective evidence retrieval from the same memory bank.
Combining both trained capabilities achieves the highest overall score of 88.80\%.
These results show that \textsc{Uni-Mem} improves both memory construction and retrieval, with each capability contributing gains even when paired with its untrained counterpart.

\begin{figure*}[t]
    \centering
    \includegraphics[width=\textwidth]{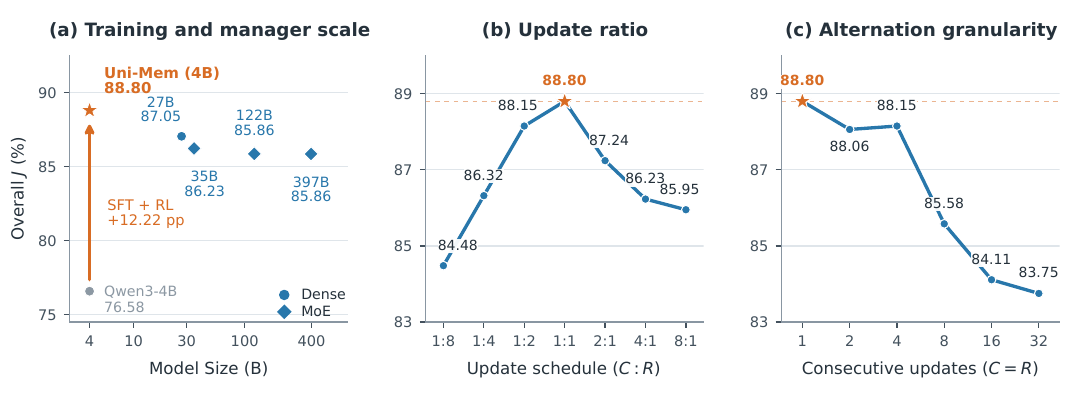}
    \vspace{-0.8cm}
    \caption{
    Effects of manager training, scale, and alternating schedules on LoCoMo.
    (a) Training and manager scale. The arrow shows the gain from SFT followed by RL.
    (b) Construction--retrieval update ratio.
    (c) Alternation granularity at a fixed $1:1$ ratio.
    $C$ and $R$ denote consecutive construction and retrieval updates per cycle.
    }
    \vspace{-0.2cm}
    \label{fig:further_analysis}
\end{figure*}

% Category-level results are provided in Appendix~\ref{app:manager_scale}.

% \paragraph{Sensitivity to the Alternating Training Schedule.}
% %claim 为什么要做alternate？
% \todo{11}
% We analyze how the construction-retrieval update ratio and alternation granularity affect Stage-Decoupled GRPO (Figure~\ref{fig:further_analysis}(b,c)).
% Here, $C$ and $R$ denote consecutive construction and retrieval updates per alternating cycle.
% A balanced $1:1$ schedule achieves the highest overall $J$ score of 88.80\%; shifting the ratio to $1:8$ or $8:1$ reduces performance by 4.32 and 2.85 percentage points, respectively.
% At a fixed $1:1$ ratio, increasing the update block length from $1:1$ to $32:32$ lowers the score to 83.75\%.
% These results indicate that both balanced allocation and frequent alternation are important for coordinated construction and retrieval learning under a shared policy. Category-level results are provided in Appendix~\ref{app:schedule_sensitivity}.

\noindent \textbf{Effect of Alternation Frequency and Update Balance.}
To determine how construction and retrieval should be
combined for joint learning within a shared policy,
we compare training schedules under Stage-Decoupled GRPO
(Figure~\ref{fig:further_analysis}(b,c)).
Each schedule $C:R$ consists of $C$ consecutive construction
updates followed by $R$ retrieval updates. As shown in Figure~\ref{fig:further_analysis}(c), extending the update blocks from $1:1$ to $32:32$ reduces
the overall $J$ score from $88.80\%$ to $83.75\%$,
indicating that less frequent alternation hinders
shared-policy learning. As shown in Figure~\ref{fig:further_analysis}(b), the $1:8$ and $8:1$ schedules reduce performance by
4.32 and 2.85 percentage points, respectively, indicating
that allocating too many updates to either stage also
harms unified learning. These results indicate that both balanced allocation and frequent alternation are important for coordinated construction and retrieval learning under a shared policy.
% Category-level results are provided in
% Appendix~\ref{app:schedule_sensitivity}.

% \noindent \textbf{Effect of Training and Manager Scale.}
% We compare \textsc{Uni-Mem} with base Qwen3-4B-Instruct and four larger off-the-shelf Qwen3.5 models, all serving as memory managers within the same agent harness.
% As shown in Figure~\ref{fig:further_analysis}(a), SFT followed by RL improves the 4B manager's overall $J$ score from 76.58\% to 88.80\%.
% \textsc{Uni-Mem} also outperforms the strongest dense model Qwen3.5-27B and MoE model Qwen3.5-397B-A17B.
% These results demonstrate that a small-size manager trained jointly for construction and retrieval can outperform substantially larger off-the-shelf alternatives. \textbf{Please refer to Appendix~\ref{app:supplementary_experiments} for more experimental results and analysis.}

\lingan{
\noindent \textbf{Effect of Training and Manager Scale.}
We compare \textsc{Uni-Mem} with the base Qwen3-4B-Instruct and four larger off-the-shelf Qwen3.5 models, all serving as memory managers within the same agent harness.
As shown in Figure~\ref{fig:further_analysis}(a), SFT followed by RL improves the 4B manager's $J$ score from 76.58\% to 88.80\%.
\textsc{Uni-Mem} also outperforms all off-the-shelf managers, including the strongest one, Qwen3.5-27B (87.05\%), and the largest one, Qwen3.5-397B-A17B (85.86\%).
These results demonstrate that a small manager trained jointly for construction and retrieval can outperform substantially larger off-the-shelf alternatives.
% \textbf{Please refer to Appendix~\ref{app:supplementary_experiments} for more experimental results and analysis.}
}

\begin{figure}[t]
    \centering
    \includegraphics[width=\columnwidth]{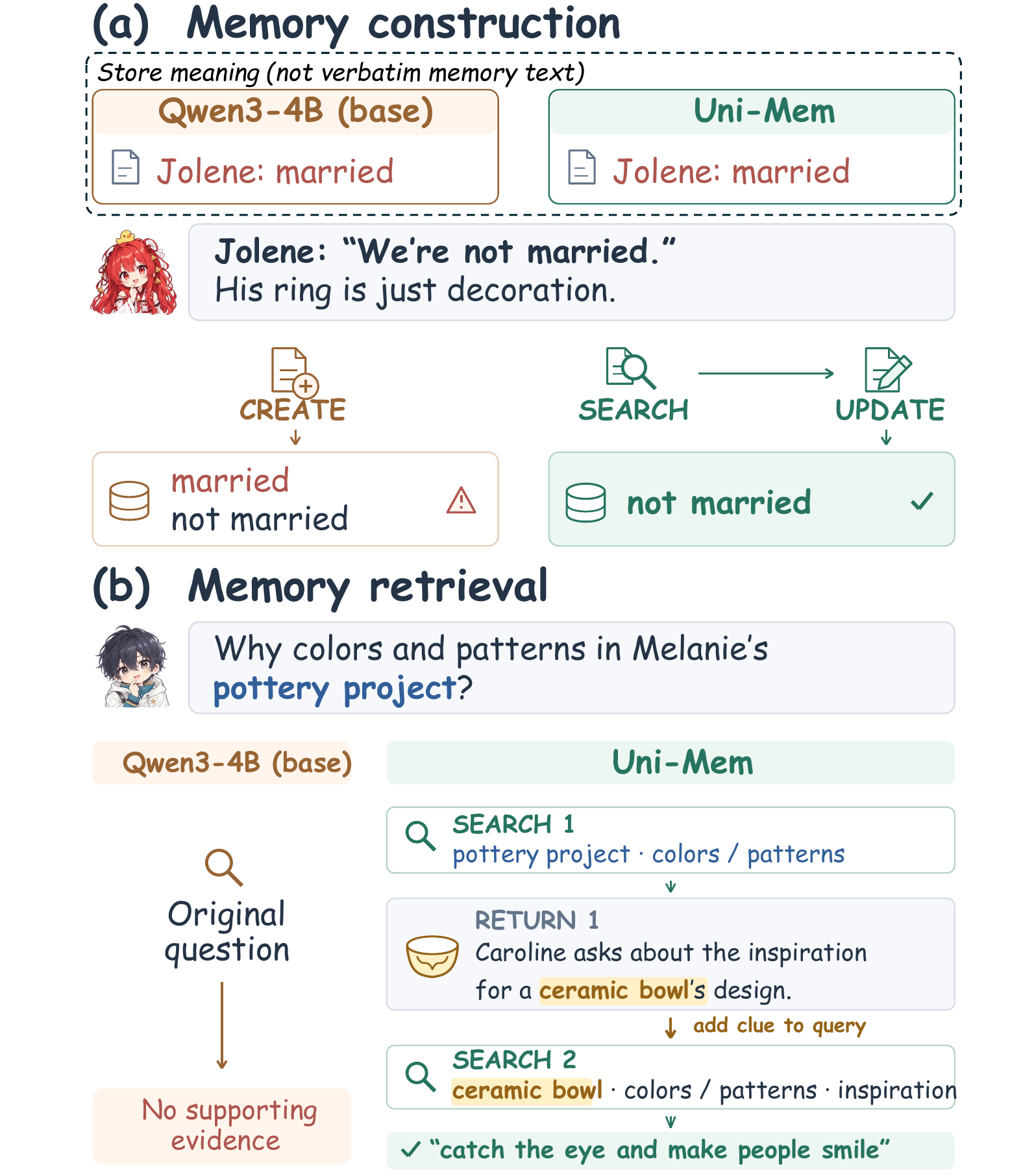}
    \caption{
    Case studies comparing \textsc{Uni-Mem} and the base Qwen3-4B manager in (a) memory construction and (b) memory retrieval.
    }
    \vspace{-0.7cm}
    \label{fig:case_study}
\end{figure}

\subsection{Case Study}
\label{sec:case_study}
% As shown in figure~\ref{fig:case_study}, we compares \textsc{Uni-Mem} with the base Qwen3-4B memory manager to reveal the capabilitied acquires through our training.
% In Figure~\ref{fig:case_study}(a), both memory banks initially encode the mistaken belief that Jolene is married, with their underlying entries differing in wording and content.
% Following Jolene's clarification, \textsc{Uni-Mem} locates the conflicting entry through \texttt{SEARCH} and corrects it through \texttt{UPDATE}, while the base manager uses \texttt{CREATE} and leaves contradictory claims unresolved.
% This illustrates \emph{how correcting existing errors keeps the memory bank clean and internally consistent.}
% In Figure~\ref{fig:case_study}(b), \textsc{Uni-Mem} uses the \emph{ceramic bowl} reference discovered during its initial search to refine the next query and retrieve the answer-bearing memory.
% The base manager searches with the original question and misses this evidence. It reveals the iterative retrieval capabilities of Uni-mem.
As shown in Figure~\ref{fig:case_study}, we compare
\textsc{Uni-Mem} with the base Qwen3-4B memory manager
to illustrate the capabilities acquired through our training.
In Figure~\ref{fig:case_study}(a), both memory banks initially encode the mistaken belief that Jolene is married, with their underlying entries differing in wording and content.
Following Jolene's clarification, \textsc{Uni-Mem} locates the conflicting entry through \texttt{SEARCH} and corrects it through \texttt{UPDATE}, while the base manager uses \texttt{CREATE} and leaves contradictory claims unresolved.
This illustrates \textsc{Uni-Mem}'s ability to
\emph{identify and resolve memory conflicts, maintaining
an internally consistent memory bank}.
In Figure~\ref{fig:case_study}(b), \textsc{Uni-Mem} uses the \emph{ceramic bowl} reference discovered during its initial search to refine the next query and retrieve the answer-bearing memory. This highlights \textsc{Uni-Mem}'s ability to
\emph{iteratively refine its search based on intermediate
results to retrieve answer-relevant evidence}.

\section*{Conclusion}
We presented \textsc{Uni-Mem}, a unified memory manager
that jointly learns agentic memory construction and
retrieval through end-to-end online reinforcement learning
under a shared policy.
By combining Stage-Decoupled GRPO with Event-Wise Credit
Assignment, \textsc{Uni-Mem} obtains stage-specific learning
signals and assigns credit to relevant memory operations,
enabling coordinated online learning of construction and
retrieval over long horizons.
Experiments demonstrate that \textsc{Uni-Mem} outperforms
strong RL-based memory baselines on LoCoMo and generalizes
to LongMemEval.
% Further analyses show improvements in both construction
% and retrieval, supporting the effectiveness of jointly
% optimizing these capabilities under a shared policy.

\section*{Limitations}
%无关重要的  end to end 训练慢，依赖judge model
%
\textsc{Uni-Mem} unifies memory construction and retrieval through online RL, enabling coordinated policy learning with stage-specific optimization and event-wise credit assignment. However, training remains resource-intensive because rewards require completing downstream QA and obtaining judge-model evaluations. These evaluations typically rely on external APIs or stronger deployed models, introducing more costs and latency. Running judge evaluations concurrently or adopting asynchronous RL could alleviate the latency bottleneck. Future work will explore more robust process-level supervision to reduce reliance on outcome-based evaluations.


\bibliography{custom}

% \clearpage
\appendix

\section*{Appendix Overview}
This appendix provides supplementary methodological details and experimental analyses to complement the main text and facilitate reproducibility, organized as follows:
Appendix~\ref{app:related_comparison} positions \textsc{Uni-Mem} relative to existing memory agents in terms of interaction patterns and training paradigms, highlighting its differences from Mem-T.
Appendix~\ref{app:implementation_details} details the memory representation, provenance tracking, tool interfaces, and execution protocol.
Appendix~\ref{app:training_details} describes supervised cold-start training, event-wise credit computation, RL hyperparameters, and computational resources.
Appendix~\ref{app:baseline} summarizes the baseline methods.
Appendix~\ref{app:experimental_setup} documents the Memory-R2 re-evaluation, judge-prompt sensitivity, and ablation configurations.
Appendix~\ref{app:supplementary_experiments} presents training curves and detailed experimental results.
Finally, Appendix~\ref{app:prompt_templates} provides the prompt templates for memory management, answer generation, and LLM-as-a-Judge evaluation.

% \todo{

% \section{To Do List}
% \begin{itemize}
%     % \item introduce the implementation details, including sft data, lr, batch size and so on.
%     \item  UPDATE calls after training. After training, the manager calls the UPDATE tool less often, and this is driven by the task. Our method remains compatible with update-aware credit assignment. Add a figure or table to show this?
%     \item Overview of UNI-MEM sign need update according to the methods
%     % \item In ablation study, how does our sigle stage form?
%     % \item  Memory-R2 reproduction gap. Explain in detail why our reproduced Memory-R2 results are much lower than the reported ones: the difference comes from the judge prompt, and running the officially released checkpoint gives exactly these results.
% \end{itemize}
% }

\section{Positioning \textsc{Uni-Mem} with Related Methods}
\label{app:related_comparison}

We position \textsc{Uni-Mem} with RL-trained memory agents from two
perspectives: how the agent interacts with tools within a single
construction or retrieval event, and how the two stages are trained.
Table~\ref{tab:related_comparison} gives an overview.

\begin{table}[h]
\centering
\footnotesize
\setlength{\tabcolsep}{3pt}
\begin{tabular}{@{}lcccc@{}}
\toprule
& \multicolumn{2}{c}{\textbf{Interaction}} & \multicolumn{2}{c}{\textbf{Training}} \\
\cmidrule(lr){2-3}\cmidrule(lr){4-5}
\textbf{Method} & \textbf{Con.} & \textbf{Ret.} & \textbf{Con.} & \textbf{Ret.} \\
\midrule
MEM1        & S & --    & RL      & --    \\
MemAgent    & S & --    & RL      & --    \\
Memory-R1   & S & Fixed & RL      & Fixed \\
Mem-$\alpha$& S & Fixed & RL      & Fixed \\
Memory-R2   & S & Fixed & RL      & Fixed \\
Mem-T       & P & M     & Offline & RL    \\
\midrule
\textsc{Uni-Mem} & M & M & RL & RL \\
\bottomrule
\end{tabular}
\caption{Interaction pattern and training of construction (Con.)
and retrieval (Ret.). S: single-turn; P: fixed multi-step pipeline;
M: multi-turn with tool observations; Fixed: similarity retrieval
without learning; --: no explicit stage. Offline: imitation of
hindsight-filtered actions.}
\label{tab:related_comparison}
\end{table}

\paragraph{Memory manager as an agent.}
Although several prior works describe their memory modules as agents,
the memory manager in most of them acts as a one-step policy: given an
input and a context prepared by a fixed procedure, it produces its
operations in a single generation, and their results are never fed
back to it. We refer to such events as single-turn. In \textsc{Uni-Mem},
the manager is instead an agent that interacts with the memory bank
over multiple turns, as in ReAct~\citep{yao2022react}. At each step, it
chooses a tool call, observes the returned result, and decides what to
do next, including when to stop.
Among prior methods, MEM1~\citep{zhou2026mem1} interacts with an
external environment over multiple turns, but its memory is an
internal state rewritten in one generation at every turn, with no tool
access to the memory itself. MemAgent~\citep{yu2026memagent} overwrites
a fixed-length memory once per chunk and answers from the final memory.
Memory-R1~\citep{yan2026memory_r1}, Memory-R2~\citep{yan2026memory_r2},
and Mem-$\alpha$~\citep{wang2025mem} generate the memory operations for
each input in one pass, and answer questions with fixed similarity
retrieval (embedding or BM25).
Mem-T~\citep{yue2026mem-t} performs multi-turn search at query time,
but its construction is a fixed two-step pipeline: a formation step
proposes memory candidates, and an evolution step decides one
operation for each candidate. Neither step issues its own searches or
sees the execution results of earlier operations.

\paragraph{Why an agentic manager helps.}
The key difference is who decides what the manager sees. In
single-turn methods, the context for each decision is prepared by a
fixed similarity search against the incoming fact or the raw query;
if it misses a relevant record, the policy cannot recover, and RL
cannot improve this step. An agentic manager gathers its own context,
making this step part of the learned policy. During construction, it
can search for records that a new chunk duplicates or contradicts
before writing. During retrieval, it can reformulate the query and
follow intermediate results for multi-hop questions, as found useful
in multi-step QA~\citep{trivedi2023ircot,jin2025searchr1}. In line
with this, the trained retrieval policy improves the overall $J$ score
by 4.50 and 4.04 points on the same memory banks
(Table~\ref{tab:construction_search}).

\subsection{Differences from Mem-T}

Mem-T is the closest work to ours. Beyond the construction pipeline
discussed above, the main difference lies in how construction is
trained. Mem-T trains retrieval online with MoT-GRPO but learns
construction offline, imitating actions that are scored in hindsight
from retrieval rollouts. These scores come from an earlier policy and
memory state and can become outdated as the shared parameters change.
Moreover, they depend on whether the sampled retrieval surfaces the
stored memory, so a correctly stored fact can receive little credit
when retrieval fails. \textsc{Uni-Mem} trains both stages with
on-policy RL and scores construction with evidence-indexed memory
scoring (Eq.~\eqref{eq:memory_scoring}), which does not depend on the
learned retriever. Consistent with this, longer alternation blocks,
which let one stage lag behind the other, lower the overall $J$ score
from 88.80\% to 83.75\% (Figure~\ref{fig:further_analysis}(c)).

% Appendix A
\section{Implementation Details}
\label{app:implementation_details}

\subsection{Memory Representation and Provenance Tracking}
\label{app:memory_representation}

\lingan{
We use the storage and retrieval layer of Mem0~\cite{chhikara2025mem0}
(with Qdrant as the vector store) as the memory bank, rather than a
naive record list, since a capable manager should be able to operate
such a well-engineered store.
Crucially, we bypass Mem0's native interface, in which a fixed LLM
pipeline extracts facts and chooses among \texttt{ADD},
\texttt{UPDATE}, \texttt{DELETE}, and \texttt{NOOP}, and the question
is issued as a single search.
Instead, \emph{the learned policy decides what to store, when and how to
revise records, and when and how to search}; Mem0 only executes these
decisions.
\texttt{CREATE} inserts agent-written content verbatim without Mem0's
fact extraction, \texttt{UPDATE} replaces a record's text and
embedding under the same ID, \texttt{DELETE} removes it, and
\texttt{SEARCH} returns the most similar records with their IDs and text.
% Each record additionally stores the index $c_i$ of the chunk whose
% construction event wrote it as metadata; this provenance is hidden from
% the manager and used only for evidence-indexed scoring during training
% (Eq.~\eqref{eq:memory_scoring}).
The single source-chunk index $c_i$ in the main text is a
simplified notation for record provenance. In implementation,
each record maintains a cumulative list of chunk IDs:
\texttt{CREATE} initializes this list with the current chunk ID,
and \texttt{UPDATE} appends the current chunk ID while retaining
previous associations.
$\operatorname{Index}(E_k,\mathcal{B})$ selects the current
versions of all surviving records whose provenance lists
intersect $E_k$, including records updated rather than
originally created in an evidence chunk.
This metadata is hidden from the manager and used only for
evidence-indexed scoring (Eq.~\eqref{eq:memory_scoring})
and credit assignment during training.
}
\subsection{Tool Specifications and Execution Protocol}
\label{app:tool_specifications}

Tools~\ref{tool:search}--\ref{tool:finish} specify the manager's
interface when using an external answer model.
Each card lists the inputs, outputs, and execution semantics.
Arguments are required unless marked optional.

\begin{toolbox}[label={tool:search}]{SEARCH}
memory_search(query, limit)

[Input]
query: string; semantic search query.
limit: integer, optional; requested result limit.

[Output]
Matching memory records with IDs and text.
The result set may be empty.

[Execution]
Search the current memory bank without modifying
it. The environment determines the effective
result limit from the event configuration.
\end{toolbox}

\begin{toolbox}[label={tool:create}]{CREATE}
create_memory(content)

[Input]
content: string; one self-contained memory.

[Output]
Created record: id, memory, and event="ADD".

[Execution]
Insert the supplied text as a new record.
Preserve names, dates, and quantities; store
each new fact in a separate record.
\end{toolbox}

\begin{toolbox}[label={tool:update}]{UPDATE}
update_memory(memory_id, content)

[Input]
memory_id: string; target record ID.
content: string; replacement memory text.

[Output]
Update confirmation, or an error on failure.

[Execution]
Replace the record's text and embedding while
preserving its ID. Use for an explicit correction
or supersession of the same stored fact.
Store new facts through CREATE.
\end{toolbox}

\begin{toolbox}[label={tool:delete}]{DELETE}
delete_memory(memory_id)

[Input]
memory_id: string; target record ID.

[Output]
Deletion confirmation, or an error on failure.

[Execution]
Remove the specified record from the memory bank.
Use when the record is wrong and no replacement
is needed.
\end{toolbox}

\begin{toolbox}[label={tool:finish}]{FINISH}
finish_event(summary)

[Input]
summary: string, optional; brief completion note.

[Output]
The next event's observation, or termination
after the final event. A rejected call returns
an error observation.

[Execution]
Close the current event. For retrieval events,
hand the accumulated evidence to the external
answer model before advancing.

[Call example]
<tool_call>
{"name": "finish_event", "arguments": {}}
</tool_call>
\end{toolbox}

\paragraph{Construction events.}
Each event starts with a chunk of up to two new dialogue messages
and up to ten preceding turns from the same session as read-only
context.
The manager may interleave searches with memory writes.
A write requires a successful search response observed in an
earlier assistant turn of the same event; an empty result is valid.
A write bundled with the first search is therefore rejected.
\texttt{FINISH} closes the event and advances to the next input,
including when no memory changes are needed.

\paragraph{Retrieval events.}
Each event starts with a question and a reference date, when available.
Only \texttt{SEARCH} and \texttt{FINISH} are permitted.
The manager may refine its queries over multiple rounds;
retrieved records are accumulated and deduplicated by ID.
A voluntary \texttt{FINISH} requires a successful search response
and hands the accumulated evidence to the frozen answer model.
The environment then clears the retrieval cache and advances
to the next question, or terminates after the final question.

\paragraph{Execution rules.}
Each call is a JSON object containing \texttt{name} and
\texttt{arguments}, enclosed in \texttt{<tool\_call>} tags.
Calls within a response execute sequentially, and their results
or errors become observations for the next decision.
For updates and deletions, displayed memory aliases are resolved
to backend record IDs.
\texttt{FINISH} must be issued alone: when bundled with other
calls, it is ignored while the remaining calls undergo
their usual checks.
In our RL configuration, each event permits at most eight
searches and 24 manager responses; exhausting the response
budget forces event closure.
The search result limit defaults to 10 for construction and
20 for retrieval. Smaller requested limits are raised to
these values, and all limits are capped at 200.

\section{Training Details}
\label{app:training_details}

\subsection{Supervised Cold-Start Training}
\label{app:supervised_cold_start}

% \paragraph{sft data 如何制作,结果规模?}

% \paragraph{base,lr,batch size ,how to select toolformat%(gate) pass@1 pass @3  gap f1 mean_entropy 在locomo的val上测试}

% %为了建立agent调用工具,管理记忆的基础能力,我们利用gpt-5.4作为教师模型去训练我们的,我们参考locomo的

% Paste under the existing Supervised Cold-Start Training subsection.
% Source-data counts are from the supplied four-conversation statistics.
% These counts describe source tasks, not the number of SFT training examples.
% Requires booktabs and graphicx, already included in the manuscript.

\paragraph{Source data construction.}
Following the data-generation methodology of LoCoMo~\cite{locomo2024evaluating},
we construct four additional multi-session conversation tasks as source data
for SFT. These tasks span short, medium, long, and extra-long histories,
ranging from 162 to 1,054 turns. As summarized in
Table~\ref{tab:sft_source_data}, the corpus contains 117 sessions,
2,290 turns, and 773 question--answer pairs covering single-hop, multi-hop,
temporal, and open-domain questions.

\begin{table*}[t]
\centering
\small
\setlength{\tabcolsep}{5pt}
\renewcommand{\arraystretch}{1.1}
\resizebox{\textwidth}{!}{%
\begin{tabular}{@{}llrrrrrrr@{}}
\toprule
& & & & & \multicolumn{4}{c}{\textbf{QA category}} \\
\cmidrule(lr){6-9}
\textbf{Length} & \textbf{Task} & \textbf{Sessions} & \textbf{Turns}
& \textbf{QA pairs} & \textbf{Single-hop} & \textbf{Multi-hop}
& \textbf{Temporal} & \textbf{Open-domain} \\
\midrule
Short & \texttt{conv\_05} & 10 & 162 & 114 & 66 & 21 & 16 & 11 \\
Medium & \texttt{conv\_07} & 20 & 371 & 128 & 68 & 23 & 25 & 12 \\
Long & \texttt{conv\_12} & 35 & 703 & 116 & 57 & 13 & 34 & 12 \\
Extra-long & \texttt{conv\_19} & 52 & 1,054 & 415 & 198 & 74 & 113 & 30 \\
\midrule
\textbf{Total} & \textbf{4 tasks} & \textbf{117} & \textbf{2,290} & \textbf{773} & \textbf{389} & \textbf{131} & \textbf{188} & \textbf{65} \\
\bottomrule
\end{tabular}%
}
\caption{Statistics of the four source tasks for SFT demonstration generation.
A turn is one nonempty message; session dates and speaker metadata are excluded.
QA counts include all source questions, with no adversarial questions.}
\label{tab:sft_source_data}
\end{table*}

\paragraph{SFT demonstration generation.}
To initialize the agent's tool-use and memory-management capabilities,
we use GPT-5.4 as a teacher to generate interactive demonstrations on these
four source tasks. The teacher processes dialogue chunks to construct
memory and performs tool-based retrieval for the associated questions,
conditioning each decision on preceding tool observations. We record the
tool calls and returned observations as supervised dialogue sequences,
with teacher responses serving as training targets.

\paragraph{Optimization and checkpoint selection.}
\begin{figure}[t]
    \centering
    \includegraphics[width=\linewidth]{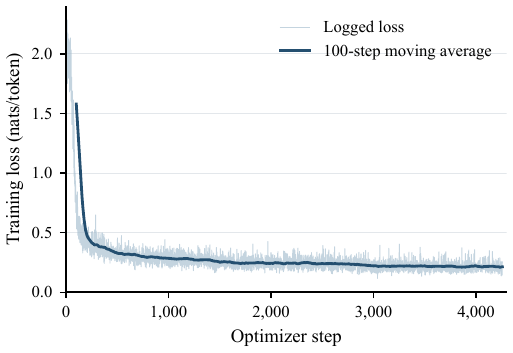}
    \caption{SFT training loss, measured as the mean negative
    log-likelihood over supervised assistant tokens (nats/token).
    The horizontal axis counts optimizer updates.
    The light curve shows the logged loss; the dark curve shows
    a trailing 100-step moving average.}
    \label{fig:sft_train_loss}
\end{figure}
We fine-tune Qwen3-4B-Instruct with a learning rate of $10^{-5}$,
a warmup ratio of 0.1, and a global batch size of 8
(one example per GPU on eight GPUs).
Training uses bfloat16 precision and a maximum sequence length of
16,384 tokens, with loss applied only to assistant tokens.
The SFT budget is set to three epochs, with checkpoint saving
and validation-loss evaluation every 431 optimizer updates.
Figure~\ref{fig:sft_train_loss} shows the recorded training loss over
4,257 optimizer updates. 
The mean loss decreases from 1.579 over the
first 100 updates to 0.213 over the last 100 updates.

For checkpoint selection, we evaluate downstream QA on the LoCoMo
validation partition from the $2{:}1{:}7$ train/validation/test split
used in the main experiments. We first apply a 95\% tool-format gate,
measuring the fraction of assistant turns with at least one parseable,
recognized tool call; this measures format compliance rather than
execution success.
Across three stochastic runs, pass@1 is the mean judge accuracy,
while pass@3 is the fraction of questions answered correctly in at
least one run. Their difference, \emph{gap}, measures the additional
coverage available through sampling. F1 provides a complementary
token-overlap diagnostic. To assess policy diversity, we also measure
mean next-token entropy on held-out SFT prompts, using the partial
entropy sum over returned top-token probabilities.
Finally, \emph{finish-bundled} counts turns that combine
\texttt{finish\_event} with another action, which the environment rejects.

Table~\ref{tab:sft_checkpoint_selection} reports the checkpoints
evaluated for downstream QA, from step 431 to step 3448
at intervals of 431 optimizer updates.
We select \textbf{step 862} for subsequent RL, balancing task performance,
tool-use behavior, and policy diversity. Compared with step 431,
pass@1 improves from 81.93\% to 83.35\%, and finish-bundled turns
decrease from 182 to 54, while the entropy estimate remains at
0.4058 nats/token and the gap at 11.06 percentage points.
Although step 1724 achieves the highest pass@1 and F1,
step 862 retains higher entropy and a larger sampling gap.

\begin{table*}[t]
\centering
\small
\setlength{\tabcolsep}{5pt}
\renewcommand{\arraystretch}{1.1}
\resizebox{\textwidth}{!}{%
\begin{tabular}{@{}lrrrrrrr@{}}
\toprule
\textbf{Checkpoint} & \textbf{Tool format (\%)} & \textbf{Pass@1 (\%)}
& \textbf{Pass@3 (\%)} & \textbf{Gap (pp)} & \textbf{F1 (\%)}
& \textbf{Entropy} & \textbf{Finish-bundled} \\
\midrule
Base (pre-SFT) & 99.32 & 79.57 & 90.45 & 10.89 & 24.80 & 0.1706 & 1,610 \\
Step 431 & 99.60 & 81.93 & 94.94 & 13.01 & 26.52 & 0.4482 & 182 \\
\textbf{Step 862 (selected)} & \textbf{99.34} & \textbf{83.35} & \textbf{94.42} & \textbf{11.06} & \textbf{26.96} & \textbf{0.4058} & \textbf{54} \\
Step 1293 & 99.70 & 83.59 & 94.09 & 10.50 & 26.98 & 0.3717 & 67 \\
Step 1724 & 99.74 & 85.19 & 94.94 & 9.74 & 27.66 & 0.3776 & 21 \\
Step 2155 & 99.81 & 85.06 & 94.55 & 9.48 & 27.24 & 0.3632 & 21 \\
Step 2586 & 99.77 & 84.72 & 94.94 & 10.22 & 27.32 & 0.3159 & 53 \\
Step 3017 & 99.84 & 84.37 & 94.55 & 10.17 & 27.46 & 0.3239 & 46 \\
Step 3448 & 99.69 & 83.57 & 95.32 & 11.75 & 27.14 & 0.3424 & 29 \\
\bottomrule
\end{tabular}%
}
\caption{SFT checkpoint selection for RL initialization.
Bold marks the selected checkpoint, not per-column maxima.
All checkpoints pass the 95\% tool-format gate.
Pass@1, pass@3, and F1 use judged, non-adversarial QA records;
gap is pass@3 minus pass@1 in percentage points (pp), computed
before rounding. Entropy is a partial-sum estimate in nats/token
on held-out SFT prompts. Tool format and finish-bundled are
measured from the first evaluation run; finish-bundled is a raw count.}
\label{tab:sft_checkpoint_selection}
\end{table*}

\subsection{Event-Wise Credit Computation}
\label{app:credit_computation}
 
% We detail the update-robust evidence attribution in
% Section~\ref{sec:method_eca}. During construction rollouts, we log the
% record modified by each \texttt{UPDATE} or \texttt{DELETE} together
% with the event that performs it, which links every record to its later
% versions. For query $q_k$, we follow the records written in
% $\mathcal{E}_k^{\mathrm{con}}$ to their current versions in
% $\mathcal{B}^{(n)}$, dropping those whose chain ends in a deletion.
% The remaining versions form $\operatorname{Index}(E_k,\mathcal{B}^{(n)})$,
% and all events along these chains are added to
% $\mathcal{E}_k^{\mathrm{con}}$. For example, if event~5 creates a
% record from an evidence chunk, event~12 updates it, and event~20
% deletes it, then $\mathcal{S}_k(\mathcal{B}^{(n)})=0$ and all three
% events receive~$0$; without this attribution, only event~5 would be
% credited. The lineage is used only for computing credit and requires
% no additional calls to the answer model.
We detail the update-robust evidence attribution in
Section~\ref{sec:method_eca}. During construction rollouts,
we log each record's \texttt{UPDATE} and \texttt{DELETE}
operations together with their triggering events.
For query $q_k$, we identify records created or updated
in the original evidence chunks $E_k$ and follow their
subsequent modifications. Their surviving current versions
form $\operatorname{Index}(E_k,\mathcal{B}^{(n)})$.
The rollout-specific expanded credit set,
$\widetilde{\mathcal{E}}_k^{\mathrm{con},(n)}$, contains the
original evidence events and subsequent update or deletion
events on these records, and is used in
Eq.~\eqref{eq:construction_credit}.
This expansion affects only credit recipients: memory
selection always uses the original $E_k$, without including
unrelated records written by the added events.
For example, if event~5 creates an evidence record,
event~12 updates it, and event~20 deletes it, all three
events share the score computed from the remaining indexed
memories. Deletion alone does not imply
$\mathcal{S}_k(\mathcal{B}^{(n)})=0$.
The modification logs remain available after deletion,
and lineage-based credit assignment requires no additional
answer-model calls.
% \subsection{Complete Training Procedure for Stage-Decoupled GRPO}
% \label{app:training_procedure}

\subsection{RL Training Hyperparameters and Computational Resources}
\label{app:training_configuration}

We optimize the SFT-initialized Qwen3-4B-Instruct manager with
Stage-Decoupled GRPO, alternating construction and retrieval
iterations at a $1{:}1$ ratio.
During RL training, Qwen3.8-27B serves as both the answer model and
the judge, using the Mem0 judge prompt, while Qwen3-Embedding-4B
provides memory embeddings. These service models remain frozen.
Table~\ref{tab:rl_training_hyperparameters} summarizes the main settings.
We use two Ascend 910B nodes: one runs RL on eight NPUs with
\texttt{verl} 0.8, and the other hosts the answer, judge, embedding,
and memory-storage services. 

\lingan{
\paragraph{Answer and judge models during training.}
RL training requires a large number of answer-model and judge calls:
every construction rollout is scored on all training queries through
evidence-indexed scoring, and every retrieval rollout is answered and
judged. Due to computational and API cost constraints, we use a
locally deployed Qwen3.8-27B as both the frozen answer model $g_\phi$
and the judge $\mathcal{J}_\psi$ during training. For evaluation, we
use GPT-OSS-120B as the answer model and GPT-4o-mini as the judge. Since the manager is trained with one
answer model and evaluated with another, the reported gains also
suggest that the learned memory behaviors transfer across answer
models.
}

\begin{table}[t]
\centering
\small
\setlength{\tabcolsep}{4pt}
\renewcommand{\arraystretch}{1.08}
\begin{tabular}{@{}lr@{}}
\toprule
\textbf{Hyperparameter} & \textbf{Value} \\
\midrule
Actor learning rate & $2\times10^{-6}$ \\
Task batch size & 2 \\
Rollouts per task ($N$) & 8 \\
PPO mini-batch size & 1 \\
Sampling temperature / top-$p$ & 0.7 / 1.0 \\
Training epochs & 200 \\
% Prompt / response token limits & 28,000 / 8,000 \\
% Maximum sampled QA pairs per task & 100 \\
Advantage normalization & Group mean/std \\
% Policy loss reduction & Token mean \\
KL / entropy coefficients & 0 / 0 \\
Model precision & bfloat16 \\
% Sequence parallelism degree & 2 \\
Validation / checkpoint interval & 10 / 10 steps \\
\bottomrule
\end{tabular}
\caption{Main RL training hyperparameters. Task batch size is
specified before sampling the eight rollouts per task.}
\label{tab:rl_training_hyperparameters}
\end{table}

% Training for 200 epochs takes approximately 10 days.

\section{Details of Baselines.}
\label{app:baseline}

We describe the baseline methods below. Training-free methods manage memory without task-specific parameter updates, whereas training-based methods learn memory-related policies from task feedback.

\paragraph{RAG.}
RAG indexes conversation history and retrieves relevant text segments through embedding similarity. The retrieved segments are provided to an answer model as supporting context. It provides a basic conversation-retrieval baseline without explicit fact extraction or memory editing.

\paragraph{A-Mem.}
A-Mem~\citep{xu2026mem} organizes interactions into structured notes containing the original content, timestamps, keywords, tags, and contextual descriptions. It links semantically related notes and revises existing notes when new information arrives. At query time, relevant notes are retrieved through embedding similarity to support answer generation.

\paragraph{Mem0.}
Mem0~\citep{chhikara2025mem0} maintains an external memory bank through an extraction-and-update pipeline. An LLM extracts salient facts from incoming interactions, retrieves related memories, and selects among \texttt{ADD}, \texttt{UPDATE}, \texttt{DELETE}, and \texttt{NOOP} to maintain the bank. Relevant memories are subsequently retrieved to answer questions.

\paragraph{MemoryOS.}
MemoryOS~\citep{kang2025memory} organizes information into short-term, mid-term, and long-term personal memory. Dialogue records are grouped into topic-based segments, with dialogue-chain FIFO updates and heat-based consolidation transferring information across levels. Retrieval combines relevant dialogue segments and pages with recent context and persistent user and agent information for response generation.

\paragraph{MEM1.}
MEM1~\citep{zhou2026mem1} integrates memory consolidation into an agent's reasoning process. At each interaction step, it combines the previous internal state with new observations to generate an updated state and the next action. Earlier interaction content is removed from the active context, and RL trains the agent to preserve task-relevant information within this bounded context.

\paragraph{MemAgent.}
MemAgent~\citep{yu2026memagent} processes long inputs sequentially while maintaining a bounded textual memory. Each new chunk triggers a rewrite of the memory, and the final answer is generated from the question and the resulting memory. Its multi-conversation RL algorithm uses final-answer rewards to optimize the sequence of memory updates and answer generation.

\paragraph{Memory-R1.}
Memory-R1~\citep{yan2026memory_r1} contains a memory manager and a separate answer agent. The manager learns to select \texttt{ADD}, \texttt{UPDATE}, \texttt{DELETE}, or \texttt{NOOP} for extracted information. Fixed similarity-based retrieval supplies candidate memories to the answer agent, which learns to filter and reason over them through memory distillation. Both components are optimized using downstream answer correctness.

\paragraph{Memory-R2.}
Memory-R2~\citep{yan2026memory_r2} trains a shared-parameter fact extractor and memory manager over chunked, multi-session conversations. Its LoGo-GRPO algorithm combines global trajectory supervision with local session rerollouts from shared intermediate memory states to improve credit assignment. A curriculum progressively increases the session horizon, while similarity-based retrieval remains fixed. The standard variant uses a separately trained Qwen2.5-7B-Instruct answer agent. We additionally evaluate a variant that pairs the Memory-R2 memory policy with GPT-OSS-120B~\citep{agarwal2025gpt} as the answer model. This variant matches the answer model used by \textsc{Uni-Mem}, enabling a comparison of the memory systems with the same downstream answer model.

\paragraph{Mem-T.}
Mem-T~\citep{yue2026mem-t} maintains working, factual, experiential, and raw memories. Memory construction follows a formation-and-evolution pipeline, while retrieval uses multiple search steps to gather evidence. Its MoT-GRPO algorithm trains retrieval through branching rollout trees and propagates answer feedback to intermediate operations. Construction is trained through offline imitation of actions selected by hindsight scores derived from retrieval outcomes and evidence alignment.

% Appendix C
\section{Experimental Setup and Evaluation Protocol}
\label{app:experimental_setup}

\subsection{Memory-R2 Re-evaluation and Judge-Prompt Sensitivity}
\label{app:memory_r2_reevaluation}

As shown in Table~\ref{tab:memory_r2_reevaluation},
our re-evaluated Memory-R2 results are lower than those
reported in the original work~\citep{yan2026memory_r2}.
These results are obtained by directly evaluating the
officially released checkpoints, without additional
training.
The discrepancy stems from a difference in the
LLM-as-a-Judge prompt. Our evaluation adopts a prompt
aligned with Mem0~\citep{chhikara2025mem0}, which differs
from the prompt used in the original Memory-R2 evaluation
and consequently changes how generated answers are judged.
To maintain consistent judging conditions, we apply this
prompt to both \textsc{Uni-Mem} and the re-evaluated baselines.
The complete template is provided in
Appendix~\ref{app:llm_judge_prompt}.
\begin{table}[t]
    \centering
    \footnotesize
    \setlength{\tabcolsep}{2.5pt}
    \renewcommand{\arraystretch}{1.05}
    \begin{tabular}{@{}lccccc@{}}
        \toprule
        \textbf{Setting}
        & \textbf{Multi}
        & \textbf{Open}
        & \textbf{Single}
        & \textbf{Temp.}
        & \textbf{Overall} \\
        \midrule
        \multicolumn{6}{@{}l}{\textit{Memory-R2}} \\
        Reported
        & 80.93 & 67.53 & 86.80 & 69.90 & 80.99 \\
        Re-evaluated
        & 69.15 & 61.29 & 78.61 & 57.38 & 71.26 \\
        \midrule
        \multicolumn{6}{@{}l}{\textit{Memory-R2 (OSS)}} \\
        Reported
        & 85.07 & 74.14 & 90.58 & 83.33 & 87.10 \\
        Re-evaluated
        & 74.13 & 70.97 & 81.15 & 74.68 & 77.87 \\
        \bottomrule
    \end{tabular}
    \caption{Reported and re-evaluated Memory-R2 results
    on LoCoMo ($J$, \%). Reported results are
    from~\citet{yan2026memory_r2}. Re-evaluated results
    are obtained from the officially released checkpoints
    using our Mem0-aligned judge prompt.}
    \label{tab:memory_r2_reevaluation}
\end{table}

\lingan{
\paragraph{Comparability of reported baseline results.}
Except for Memory-R2, Memory-R2 (OSS), and Mem-T, which we re-evaluate
from officially released checkpoints, the baseline results in
Table~\ref{tab:performance_comparison} are taken from
\citet{yan2026memory_r2}. They were obtained under the original
evaluation protocol of that work, which differs from ours in the judge
prompt and, for some methods, the answer model.
As Table~\ref{tab:memory_r2_reevaluation} shows, the original judge
prompt yields higher scores than our Mem0-aligned prompt, by 9.73 and
9.23 overall points for Memory-R2 and Memory-R2 (OSS), respectively.
These reported results should therefore be regarded as reference
points rather than controlled comparisons. Our conclusions rely only
on the re-evaluated baselines, which share the same test split, judge
prompt, and judge model with \textsc{Uni-Mem}.
}

% \subsection{Datasets, Splits, and Preprocessing}
% \label{app:datasets}

% \subsection{Evaluation Metrics and Scoring Protocol}
% \label{app:evaluation_protocol}

% \subsection{Baseline Configurations and Re-evaluation}
% \label{app:baseline_configurations}

\subsection{Single-Stage GRPO and Ablation Configurations}
\label{app:ablation_configurations}

$M_1$ and $M_2$ sample continuous trajectories in which memory construction
is followed by retrieval. Each rollout constructs its own memory bank
$\mathcal{B}^{(n)}$ and retrieves evidence from that bank.
Both event types contribute to the same policy update.

\paragraph{$M_1$: Standard GRPO.}
QA outcome rewards are aggregated into a trajectory-level return.
Standard GRPO normalizes these returns across sibling rollouts
and broadcasts the resulting advantage to every manager decision
step, covering both construction and retrieval.

\paragraph{$M_2$: Single-stage GRPO with ECA.}
Evidence-to-event localization
(Eq.~\eqref{eq:evidence_event_localization}) first identifies
the construction events $\mathcal{E}_k^{\mathrm{con}}$ associated
with each query.
For an incorrect answer, we re-answer the query from evidence-indexed
memories and obtain
$s_k^{(n)}=\mathcal{S}_k(\mathcal{B}^{(n)})$
using Eq.~\eqref{eq:memory_scoring}.
Let $c_k^{(n)}$ denote the credit assigned to each localized
construction event and $r_k^{(n)}$ the retrieval-event credit:
\begin{equation}
(c_k^{(n)},r_k^{(n)})=
\begin{cases}
(1,1), & R_k^{(n)}=1,\\
(0,\bot), & R_k^{(n)}=0,\ s_k^{(n)}=0,\\
(1,0), & R_k^{(n)}=0,\ s_k^{(n)}=1.
\end{cases}
\end{equation}
Thus, a failed evidence-indexed re-answer attributes the failure
to construction and excludes the retrieval event.
A successful re-answer credits construction and assigns zero
reward to the failed retrieval.
Here, $\bot$ excludes an event from both group statistics and
policy optimization; it is distinct from a zero reward.

Each construction event sums the credits of its related queries;
events without related queries are excluded.
The retained returns are normalized across corresponding events
in the rollout group using Eq.~\eqref{eq:event_advantage}.
Each event's advantage is then broadcast to all manager decision
steps within that event.

% Appendix D
\section{Supplementary Experiments}
\label{app:supplementary_experiments}

\lingan{

\subsection{Gains from the Harness versus the RL Algorithm}
\label{app:gain_decomposition}

\begin{table}[t]
\centering
\footnotesize
\setlength{\tabcolsep}{6pt}
\renewcommand{\arraystretch}{1.05}
\begin{tabular}{@{}lrr@{}}
\toprule
\textbf{Setting} & \textbf{Overall $J$} & \textbf{$\Delta$ vs.\ $M_0$} \\
\midrule
\multicolumn{3}{@{}l}{\textit{Re-evaluated baselines}} \\
Memory-R2 (OSS) & 77.87 & -- \\
Mem-T & 76.49 & -- \\
\midrule
\multicolumn{3}{@{}l}{\textit{Our harness, off-the-shelf manager}} \\
Qwen3.5-27B (strongest) & 87.05 & -- \\
Qwen3.5-397B-A17B (largest) & 85.86 & -- \\
\midrule
\multicolumn{3}{@{}l}{\textit{Our harness, Qwen3-4B-Instruct w/o SFT}} \\
Start point ($M_0$) & 76.58 & -- \\
\quad + standard GRPO ($M_1$) & 76.95 & +0.37 \\
\quad + our RL ($M_4$) & 80.99 & \textbf{+4.41} \\
\midrule
\multicolumn{3}{@{}l}{\textit{Our harness, Qwen3-4B-Instruct w/ SFT}} \\
Start point ($M_0$) & 83.56 & -- \\
\quad + standard GRPO ($M_1$) & 84.02 & +0.46 \\
\quad + our RL ($M_4$) & 88.80 & \textbf{+5.24} \\
\bottomrule
\end{tabular}
\caption{Overall $J$ (\%) on LoCoMo under the same harness.
$\Delta$ is measured against the start point $M_0$ of each block.
Our RL combines ECA, Stage-Decoupled GRPO, and update-robust
attribution; see Table~\ref{tab:ablation} for per-component results
and Table~\ref{tab:manager_scale} for all off-the-shelf managers.}
\label{tab:gain_decomposition}
\end{table}

Table~\ref{tab:gain_decomposition} separates the contributions of the
agentic harness, the initialization, and the RL algorithm. First, the
harness alone does not explain the advantage over prior methods: the
untrained manager performs on par with the re-evaluated baselines.
Second, the improvement stems from our RL algorithm rather than from
RL per se: under both initializations, standard GRPO in the same
harness yields only marginal gains, whereas our RL adds 4.41 and 5.24
points, respectively. Without any SFT, our RL alone already surpasses
all re-evaluated baselines. Third, SFT and our RL are complementary:
SFT provides a stronger starting point, and our RL still adds 5.24
points on top of it despite the smaller remaining headroom. Finally,
no off-the-shelf manager in the same harness, including the strongest
(Qwen3.5-27B) and the largest (Qwen3.5-397B-A17B, a MoE model with
17B active parameters), reaches the trained 4B manager.

}

\subsection{Training Curves}
\label{app:training_curves}

\begin{figure}[t]
    \centering
    \includegraphics[width=\linewidth]{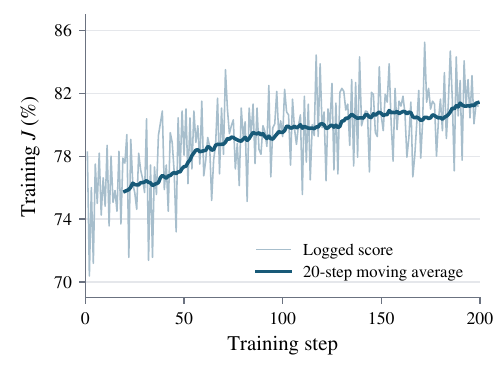}
    \caption{Training-time LLM-as-a-Judge scores ($J$, \%) of
    \textsc{Uni-Mem}, with Qwen3.8-27B used for both answer generation
    and judging. The light curve shows the logged scores;
    the dark curve shows a trailing 20-step moving average.}
    \label{fig:training_curve}
\end{figure}

Figure~\ref{fig:training_curve} reports the training-time
LLM-as-a-Judge scores ($J$) of \textsc{Uni-Mem} over 200 logged
training steps. In this run, Qwen3.8-27B serves as both
the frozen answer model and the fixed judge model, while
the Qwen3-4B-Instruct memory manager is optimized with
Stage-Decoupled GRPO. The mean $J$ increases from
$75.76\%$ over the first 20 steps to $81.41\%$ over the
final 20 steps, an improvement of $5.65$ percentage points.
Despite fluctuations in individual scores, the moving
average exhibits an overall upward trend, with smaller
changes in the later part of training.

% \subsection{Tool-Use Dynamics During Training}
% \label{app:tool_use_dynamics}

% To track changes in tool use during RL, we evaluate the manager
% every 10 training steps on the LoCoMo validation partition of
% the $2{:}1{:}7$ train/validation/test split.
% At each checkpoint, we average the logged tool-call counts over
% four validation runs, recording \texttt{SEARCH}, \texttt{CREATE},
% \texttt{UPDATE}, and \texttt{DELETE} during construction and
% \texttt{SEARCH} during retrieval.
% Figures~\ref{fig:tool_use_dynamics} and~\ref{fig:retrieval_tool_use_dynamics}
% show an overall increase in
% construction \texttt{SEARCH} and \texttt{CREATE} calls and a sustained
% rise in retrieval \texttt{SEARCH} calls.
% \texttt{UPDATE} remains infrequent, and \texttt{DELETE} is rarely used.

% \begin{figure}[t]
%     \centering
%     \includegraphics[width=\columnwidth]{picture/unimem_tool_use_dynamics.pdf}
%     \caption{Construction tool use on the LoCoMo validation set during RL
%     training. Each point is the mean logged call count over four
%     validation runs. Panels use different y-axis scales.}
%     \label{fig:tool_use_dynamics}
% \end{figure}

% \begin{figure}[t]
%     \centering
%     \includegraphics[width=\columnwidth]{picture/unimem_retrieval_tool_use.pdf}
%     \caption{Retrieval \texttt{SEARCH} calls on the LoCoMo validation
%     set during RL training, averaged over four validation runs
%     per checkpoint.}
%     \label{fig:retrieval_tool_use_dynamics}
% \end{figure}

\subsection{Detailed Results on Training and Manager Scale}
\label{app:manager_scale}

Table~\ref{tab:manager_scale} provides the category-level
results corresponding to
Figure~\ref{fig:further_analysis}(a).
We evaluate Qwen3-4B and four larger Qwen3.5 models as
off-the-shelf memory managers within the same agentic
framework.
\textsc{Uni-Mem} uses Qwen3-4B initialized through supervised
fine-tuning and subsequently optimized with Stage-Decoupled
GRPO.
All settings follow the evaluation protocol described
in the main text.

\begin{table}[t]
    \centering
    \footnotesize
    \setlength{\tabcolsep}{2.5pt}
    \renewcommand{\arraystretch}{1.08}
    \resizebox{\columnwidth}{!}{%
    \begin{tabular}{@{}lccccc@{}}
        \toprule
        \textbf{Manager}
        & \textbf{Multi}
        & \textbf{Open}
        & \textbf{Single}
        & \textbf{Temp.}
        & \textbf{Overall} \\
        \midrule
        \multicolumn{6}{@{}l}{\textit{Off-the-shelf managers}} \\
        Qwen3-4B
        & 78.61 & 70.97 & 84.04 & 57.81 & 76.58 \\
        Qwen3.5-27B
        & 85.07 & 74.19 & 89.13 & 86.92 & 87.05 \\
        Qwen3.5-35B-A3B
        & 82.09 & 74.19 & 89.30 & 85.23 & 86.23 \\
        Qwen3.5-122B-A10B
        & 84.08 & 67.74 & 88.46 & 85.65 & 85.86 \\
        Qwen3.5-397B-A17B
        & 84.08 & 75.81 & 86.93 & \textbf{87.34} & 85.86 \\
        \midrule
        \multicolumn{6}{@{}l}{\textit{Trained manager}} \\
        \textsc{Uni-Mem} (Qwen3-4B)
        & \textbf{87.06}
        & \textbf{79.03}
        & \textbf{91.51}
        & 86.08
        & \textbf{88.80} \\
        \bottomrule
    \end{tabular}%
    }
    \caption{Detailed results on training and manager scale
    on LoCoMo ($J$, \%). Multi, Open, Single, and Temp.
    denote multi-hop, open-domain, single-hop, and temporal
    questions, respectively. Best scores in each column
    are bold.}
    \label{tab:manager_scale}
\end{table}

\subsection{Detailed Analysis of Alternating Training Schedules}
\label{app:schedule_sensitivity}

Table~\ref{tab:schedule_sensitivity} provides the
category-level results corresponding to
Figure~\ref{fig:further_analysis}(b,c).
Each schedule $C{:}R$ performs $C$ consecutive construction
updates followed by $R$ consecutive retrieval updates
under Stage-Decoupled GRPO.
The first block varies the allocation of updates between
the two stages.
The second block varies the number of consecutive updates
per stage while keeping construction and retrieval updates
equally allocated.

\begin{table}[t]
    \centering
    \footnotesize
    \setlength{\tabcolsep}{3pt}
    \renewcommand{\arraystretch}{1.08}
    \begin{tabular}{@{}lccccc@{}}
        \toprule
        \textbf{Schedule}
        & \textbf{Multi}
        & \textbf{Open}
        & \textbf{Single}
        & \textbf{Temp.}
        & \textbf{Overall} \\
        \midrule
        \multicolumn{6}{@{}l}{
            \textit{Construction--retrieval update ratio}
        } \\
        $1{:}8$
        & 81.09 & 66.13 & 89.47 & 79.75 & 84.48 \\
        $1{:}4$
        & 86.07 & 64.52 & 90.83 & 81.01 & 86.32 \\
        $1{:}2$
        & 87.56 & 70.97 & \textbf{91.85} & 83.97 & 88.15 \\
        $1{:}1$
        & 87.06 & \textbf{79.03} & 91.51
        & \textbf{86.08} & \textbf{88.80} \\
        $2{:}1$
        & \textbf{89.55} & 74.19 & 89.81 & 82.28 & 87.24 \\
        $4{:}1$
        & 83.58 & 69.35 & 91.17 & 80.59 & 86.23 \\
        $8{:}1$
        & 84.58 & 72.58 & 90.15 & 80.17 & 85.95 \\
        \midrule
        \multicolumn{6}{@{}l}{
            \textit{Update block length with equal allocation}
        } \\
        $1{:}1$
        & 87.06 & 79.03 & \textbf{91.51}
        & \textbf{86.08} & \textbf{88.80} \\
        $2{:}2$
        & 88.06 & 75.81 & 90.66 & 84.81 & 88.06 \\
        $4{:}4$
        & \textbf{88.56} & \textbf{82.26}
        & 90.15 & 84.39 & 88.15 \\
        $8{:}8$
        & 83.08 & 66.13 & 89.13 & 83.97 & 85.58 \\
        $16{:}16$
        & 81.59 & 61.29 & 88.12 & 82.28 & 84.11 \\
        $32{:}32$
        & 80.10 & 62.90 & 88.62 & 80.17 & 83.75 \\
        \bottomrule
    \end{tabular}
    \caption{Detailed results for alternating training
    schedules on LoCoMo ($J$, \%).
    $C{:}R$ specifies consecutive construction and retrieval
    updates per cycle. Best scores in each column within
    each block are bold.}
    \label{tab:schedule_sensitivity}
\end{table}

\subsection{Retrieval Budget and Token Efficiency}
\label{app:retrieval_efficiency}
\begin{figure}[t]
    \centering
    \includegraphics[width=\columnwidth]{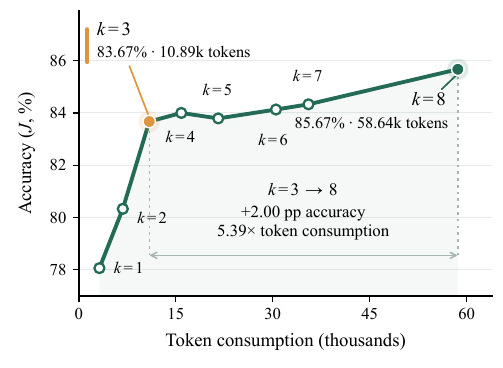}
    \caption{Accuracy ($J$) versus token consumption under different
    maximum retrieval steps $k$. Highlighted points and annotations
    compare $k=3$ and $k=8$; lines connect the measured settings.}
    \label{fig:retrieval_efficiency}
\end{figure}

To examine the accuracy-cost trade-off of iterative retrieval,
we vary the maximum retrieval steps $k$ and report $J$ and token
consumption. \textsc{Uni-Mem} serves as the memory manager, with
Qwen3.8-27B used for both answer generation and judging.
As shown in Figure~\ref{fig:retrieval_efficiency}, increasing $k$
from 1 to 3 raises $J$ from $78.06\%$ to $83.67\%$, highlighting
the benefit of multi-step retrieval guided by intermediate
evidence. Increasing $k$ from 3 to 8 yields a
further $2.00$-point gain, reaching $85.67\%$, but raises token
consumption from 10,889 to 58,642 ($5.39\times$), indicating
diminishing returns from larger retrieval budgets.

% Appendix E
% \section{Qualitative Analysis}
% \label{app:qualitative_analysis}

% \subsection{Long-Horizon Memory Construction and Retrieval}
% \label{app:memory_management_case}

% \subsection{Mechanistic Analysis of Event-Wise Credit Assignment}
% \label{app:eca_case_study}

\section{Prompt Templates}
\label{app:prompt_templates}

\subsection{Memory Manager Prompt}
\label{app:memory_manager_prompt}

The same system prompt governs memory construction and query-time retrieval
(Prompt~\ref{prompt:memory_manager}). The manager maintains long-term memory
and retrieves evidence for an external answer model through a shared tool
interface.

% Source: rl4memory_next/harness/versions/v4_3/prompts.py
% Shared SYSTEM_PROMPT_EXTERNAL; only typographic dashes are normalized.

\begin{promptbox}[label={prompt:memory_manager}]{Memory Manager Prompt}
# Role

You are an autonomous memory manager controlling a long-term memory store for
an external assistant. You see one external event at a time. The event is either
new conversation content to ingest or a user query to answer. You cannot know
how many future events remain, and you must never try to end the whole episode.

# Operating Principles

Maintain an accurate, useful memory store.

- Store durable user-specific facts, preferences, relationships, plans,
  commitments, recurring projects, explicit instructions, and dated events.
- Update a memory only when the current messages explicitly correct or
  supersede the same stored event, preference, or relationship, making its
  claim wrong or stale -- never just to add more detail. A different event, a
  different date, a compatible preference or relationship, or an unrelated
  topic is a new memory; create it.
- Delete memories only when they are wrong and no replacement is needed.
- Use the shortest useful action sequence that satisfies the current event.

# Memory Writing Standard

The memories you store are later semantically searched and handed to a separate
answerer that uses them to answer future questions. Write each memory so that
the right one is easy to retrieve by a natural-language question and complete
enough to answer from on its own.

- One focused unit per memory: a single event, preference, decision, or
  relationship. Capture that one unit with its own essential specifics
  (names, date).
- One occasion or topic per memory. Different events, different dates, or
  unrelated topics each get their own memory; never merge them into one. A memory
  that spans several dates or several unrelated events is too broad: it is hard to
  retrieve, and its single timestamp is wrong for most of what it contains.
- Concise but complete: a focused memory is usually 1-2 sentences (about 15-80
  words; up to 3 sentences / 100 words only when it is dense with proper nouns,
  dates, or quantities). The range is a guide, not a hard cap, and completeness
  always wins over word count -- never drop a proper noun, date, or number to stay
  short. When a topic carries many details, split it into several focused
  memories rather than packing one wide memory; when unsure, prefer two memories
  over one merged.
- Self-contained: understandable without other memories. Start with the
  speaker's actual name and replace pronouns with names.
- Preserve specifics verbatim: proper nouns and titles (people, places, books,
  films, brands, dishes), exact quantities, and qualifiers. Never generalize a
  concrete detail ("Ferrari 488 GTB", not "a sports car") -- the exact wording is
  what a future question is scored against.
- Anchor time to the Observation Date as absolute dates; never store
  "last week"/"recently", and never blur an explicit number or duration.
- Resolve a relative day-of-week reference to its absolute date before storing.
  If on 11 May, 2022 the conversation mentions something that happened "on
  Monday", infer the actual date (9 May, 2022) and write the memory as "On 9
  May, 2022, <what happened>" -- never store the bare weekday.

# Memory Write Gate

create_memory, update_memory, and delete_memory require previously observed
retrieval evidence for the related person, entity, date, or topic.

- Before any write tool call, you must have already observed relevant results
  from memory_search in an earlier assistant response during the
  current event.
- A retrieval result from the same assistant response is not yet observed. Do
  not call memory_search and then write in that same response.
- If you have not observed relevant retrieval evidence, do not write. Either
  retrieve first or finish the event if there is no durable memory work to do.
- Search queries should use the relevant names, entities, dates, and topic
  keywords needed to check for duplicates, stale memories, or contradictions.
- Create a new memory for each new fact, event, preference, feeling,
  reflection, question asked, or decision.
- Call update_memory only when the current messages explicitly correct or
  supersede the same retrieved event, preference, or relationship, making its
  stored claim wrong or stale and requiring in-place correction.
- Capture every fact the conversation states; do not skip one because it seems
  minor.

# Query Handling

For query events, answer from long-term memory evidence when the question
depends on remembered user-specific information.

- Retrieve and observe enough relevant memories before answering a memory-based
  query.
- If the available memories do not support an answer, do not invent one.
- You do not answer the question yourself. Retrieve useful evidence, then
  finish the event so a separate external answerer can respond from it.

# Tool Contract

You may use memory_search, create_memory, update_memory,
delete_memory, and finish_event when they are
available in the tool schema.

finish_event must be the only tool call in its assistant response. Do not bundle
finish_event with memory_search, create_memory, update_memory, or
delete_memory.

# Output Format

Wrap every tool call in:
<tool_call>
{"name": "<tool>", "arguments": {<json>}}
</tool_call>
\end{promptbox}

\subsection{LLM-as-a-Judge ($J$) Prompt}
\label{app:llm_judge_prompt}

Prompt~\ref{prompt:llm_judge} gives the \texttt{mem0\_base} evaluation
template, supplied as a user message. Given a question, a reference answer,
and a generated answer, the judge returns \texttt{CORRECT} or \texttt{WRONG}.
We report $J$ as the percentage of evaluated answers labeled
\texttt{CORRECT}. Braced fields denote runtime inputs.

% Source: rl4memory_next/eval/llm_judge.py::_JUDGE_PROMPT_MEM0_BASE
% Selected with --judge-mode mem0_base; no dialogue-evidence field is supplied.
% {answer} is the reference answer; {response} is the generated answer.
% Prompt wording is preserved; typographic quotation marks are normalized.

\begin{promptbox}[label={prompt:llm_judge}]{LLM-as-a-Judge ($J$)}
Your task is to label an answer to a question as "CORRECT" or "WRONG". You will be given the following data: (1) a question (posed by one user to another user), (2) a 'gold' (ground truth) answer, (3) a generated answer which you will score as CORRECT/WRONG.

The point of the question is to ask about something one user should know about the other user based on their prior conversations. The gold answer will usually be a concise and short answer that includes the referenced topic, for example:
Question: Do you remember what I got the last time I went to Hawaii?
Gold answer: A shell necklace
The generated answer might be much longer, but you should be generous with your grading - as long as it touches on the same topic as the gold answer, it should be counted as CORRECT.

For time related questions, the gold answer will be a specific date, month, year, etc. The generated answer might be much longer or use relative time references (like 'last Tuesday' or 'next month'), but you should be generous with your grading - as long as it refers to the same date or time period as the gold answer, it should be counted as CORRECT. Even if the format differs (e.g., 'May 7th' vs '7 May'), consider it CORRECT if it's the same date.

Now it's time for the real question:
Question: {question}
Gold answer: {answer}
Generated answer: {response}

First, provide a short (one sentence) explanation of your reasoning, then finish with CORRECT or WRONG. Do NOT include both CORRECT and WRONG in your response, or it will break the evaluation script.

Just return the label CORRECT or WRONG in a json format with the key as "label".
\end{promptbox}

\subsection{Answer Prompt}
\label{app:answer_prompt}

Prompt~\ref{prompt:answer} shows the default prompt used by the frozen
external answer model in \texttt{run\_eval}. The system message specifies
how to use memory evidence, and the user message supplies the question and
retrieved memories. Braced fields denote runtime inputs; the memory block
lists records chronologically with date prefixes. Message-role headings
are shown for clarity and are not part of the prompt text.

% Sources: rl4memory_next/prompts/answerer.py::ANSWERER_SYSTEM_PROMPT
% and rl4memory_next/utils/external_answerer.py::_format_user_prompt.
% Selection: --answerer-prompt default.
% Default strict_unknown_fallback=False: the optional fallback clause is empty.
% {reference_date} is supplied at runtime; an empty value falls back to "2023".
% {memory_block} includes the chronological-order header and dated memory lines,
% or "(No relevant memories found)" when the memory list is empty.
% Missing memory timestamps are rendered as "(unknown date)".

\begin{promptbox}[label={prompt:answer}]{Answer generation}
[System message]
You are answering a question using retrieved memories written by an agent that observed past conversations. Follow these rules internally:

Read every memory, verify the entity, combine related facts, prefer specific evidence, handle dates relative to {reference_date}, and answer directly.

Never invent specific names, titles, places, or dates that do not appear in the memories. If no memory is remotely related, use the configured fallback.

Output only:
ANSWER: <final answer>

[User message]
Question:
{question}

Retrieved memories:
{memory_block}

ANSWER:
\end{promptbox}

% Appendix B

\end{document}